\documentclass[11pt]{article}

\usepackage[final]{acl}

\usepackage{times}
\usepackage{latexsym}
\usepackage{algorithm}
\usepackage{algorithmic}
\usepackage{amssymb}
\usepackage{amsmath}
\usepackage{multicol}
\usepackage{multirow}
\usepackage{xcolor}
\usepackage{booktabs}
\usepackage{colortbl}
\usepackage{subcaption}

\usepackage[T1]{fontenc}

\usepackage[utf8]{inputenc}

\usepackage{microtype}

\usepackage{inconsolata}

\usepackage{graphicx}

\newcommand{\ourmethod}{Table-R1}
\title{Table-R1:Region-based Reinforcement Learning for Table Understanding}

\author{
  \textbf{Zhenhe Wu\textsuperscript{1,2}\footnotemark[2]},
  \textbf{Jian Yang\textsuperscript{1}\footnotemark[1]},
  \textbf{Zhongjiang He\textsuperscript{2}},
  \textbf{Changzai Pan\textsuperscript{2}},
  \textbf{Jie Zhang\textsuperscript{2}},
  \textbf{Jiaheng Liu\textsuperscript{1}},
  \\
  \textbf{Xianjie Wu\textsuperscript{1}},
  \textbf{Yu Zhao\textsuperscript{2}},
  \textbf{Shuangyong Song\textsuperscript{2}},
  \textbf{Yongxiang Li\textsuperscript{2}},
  \textbf{Zhoujun Li\textsuperscript{1}},
  \textbf{Xuelong Li\textsuperscript{3}\footnotemark[1]},
\\ 
  \textsuperscript{1}Beihang University,\\
  \textsuperscript{2}Xingchen AGI Lab, China Telecom Artificial Intelligence Technology (Beijing) Co., Ltd,\\
  \textsuperscript{3}Institute of Artificial Intelligence (TeleAI), China Telecom,
\\
\texttt{\{wuzhenhe,jiaya\}@buaa.edu.cn}
\\
\texttt{xuelong\_li@ieee.org}
}

\begin{document}
\maketitle
\footnotetext{$^*$Corresponding author.}
\footnotetext{$^\dagger$Work done during the internship at China Telecom Artificial Intelligence Technology (Beijing) Co., Ltd.}
\begin{abstract}
Tables present unique challenges for language models due to their structured row-column interactions, necessitating specialized approaches for effective comprehension. While large language models (LLMs) have demonstrated potential in table reasoning through prompting and techniques like chain-of-thought (CoT) and program-of-thought (PoT), optimizing their performance for table question answering remains underexplored. In this paper, we introduce region-based \ourmethod{}, a novel reinforcement learning approach that enhances LLM table understanding by integrating region evidence into reasoning steps. We employ Region-Enhanced Supervised Fine-Tuning (RE-SFT) to guide models in identifying relevant table regions before generating answers, incorporating textual, symbolic, and program-based reasoning. Additionally, Table-Aware Group Relative Policy Optimization (TARPO) introduces a mixed reward system to dynamically balance region accuracy and answer correctness, with decaying region rewards and consistency penalties to align reasoning steps. Experiments show that \ourmethod{} achieves an average performance improvement of 14.36 points across multiple base models on three benchmark datasets, while TARPO significantly reduces the reasoning token consumption by 67.5\% compared to GRPO, significantly advancing LLM capabilities in efficient tabular reasoning.
\end{abstract}

\section{Introduction}
Tables are a widely used data format different from plain text, which poses unique challenges for language models due to their structured row-column interactions~\citep{chain_of_table,table_meet_llm}. Understanding tabular data is crucial for applications like fact verification and question answering, driving significant research interest. Unlike plain text, tables rely on complex row-column interactions, making them challenging for models to interpret. Previous researchers have developed specialized embedding layers, attention mechanisms, and pre-training objectives to enhance structural awareness.

The rise of large language models (LLMs), such as general LLMs~\citep{gpt4,gpt45,gpt4o,telechat,wang-etal-2024-telechat,telechat2,telechat3}, has introduced new opportunities for table understanding, as their massive pre-training enables strong performance through prompting alone. Techniques like chain-of-thought~\citep{cot} (CoT) and program-of-thought~\citep{pot} (PoT) have further enhanced the reliability of LLMs by equipping responses with reasoning steps. The reasoning LLMs, such as o1/o3~\citep{deepseek_r1,o1}, introduce the step-by-step reasoning trajectories to boost test-time accuracy, which is optimized by reinforcement learning (RL). Deepseek-R1~\citep{deepseek_r1} combines group relative policy optimization~\citep{grpo} (GRPO) and a rule-based reward system to effectively improve the performance of coding and other tasks. Besides, a comprehensive and complex table question answering benchmark TableBench~\citep{tablebench} is proposed to evaluate table reasoning capabilities of LLMs.
However, prior studies often relied on general LLM capabilities and standard training paradigms without proposing training methods specifically tailored to the unique structural constraints of tables.

\begin{figure*}[tp]
\centerline{\includegraphics[width=1.88\columnwidth]{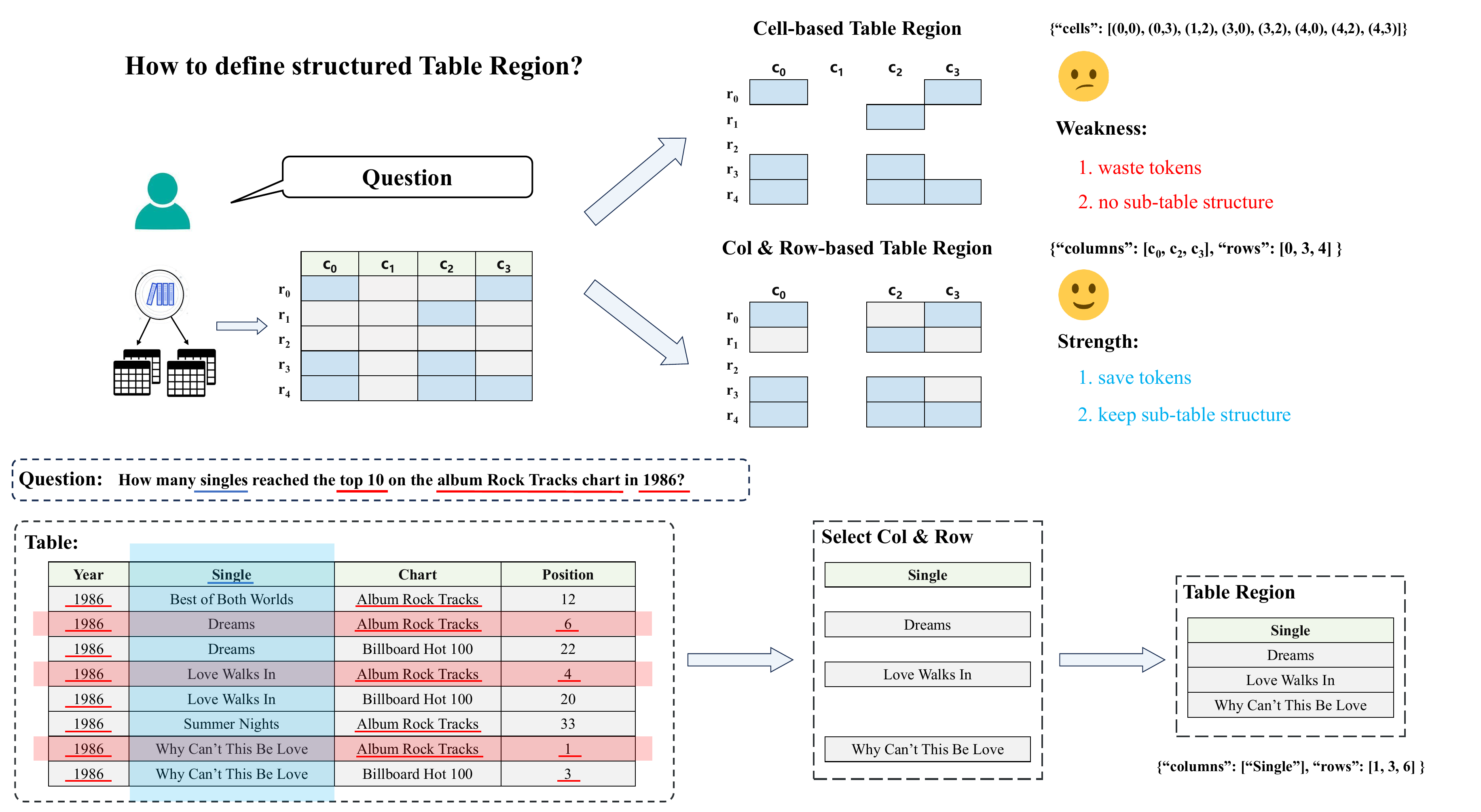}}
\caption{In \ourmethod{}, we adopt the col \& row-based table region for its structured definition. Compared to the cell-based Table Region, it not only saves input tokens but also preserves the sub-table structure. The specific example demonstrates how the LLM can extract Table Region.}
\label{table region}
\end{figure*}

To unleash the potential of LLMs in table understanding, we explore the approach of rule-based reinforcement learning to significantly strengthen table understanding capabilities by introducing the table structure information~\citep{DBLP:conf/naacl/NahidR24,DBLP:conf/sigir/YeHYLHL23,DBLP:conf/iclr/ChengX0LNHXROZS23}. We propose a region-based table question answering LLM (\ourmethod{}) by injecting the region evidence into the reasoning process and then facilitating the LLM to infer the correct answer through reinforcement learning. Specifically, \ourmethod{} is first trained with the RE-SFT (Region-Enhanced Supervised Fine-Tuning) and then enhanced by TARPO (Table-Aware Group Relative Policy Optimization). RE-SFT enhances supervised fine-tuning by guiding LLMs to first identify relevant table regions before generating answers, integrating this step into three reasoning processes, including textual chain-of-thought (TCoT), symbolic chain-of-thought (SCoT), and program-of-thoughts (PoT). TARPO extends reinforcement learning by introducing a mixed reward system that dynamically balances table region accuracy and answer correctness, with a consistency preference to ensure alignment between region identification and answer generation. By decaying the region reward weight over time and penalizing optimization inconsistencies, \ourmethod{} effectively optimizes both region identification and answer quality, significantly improving table understanding performance.

\ourmethod{} gets the top-tier performance on different benchmarks by a large margin, which demonstrates the effectiveness of our region-based table reasoning method. Our contributions and findings can be summarized as follows:

\begin{enumerate}
\item We propose a specialized LLM reinforcement training approach designed to exploits tabular structures, integrating table regions into reasoning trajectories and designing reward functions based on both table region predictions and final answers correctness.

\item We present \ourmethod{}, a unified framework that integrates RE-SFT and TARPO to synergistically guide LLMs in table region selection and answer generation across both supervised fine-tuning and reinforcement learning phases.

\item Our experimental results show that the \ourmethod{} framework delivers significant improvements in performance and generalization across various benchmarks, achieving an average performance gain of 14.36 points while reducing the reasoning tokens by 67.5\%. Furthermore, we introduce the TableInstruct-RE dataset, which enhances Chain-of-Thought (CoT) reasoning with explicit table region annotations derived from TableInstruct.
\end{enumerate}

\section{Method}
\label{sec:method}

\begin{figure*}[tp]
\centerline{\includegraphics[width=1.9\columnwidth]{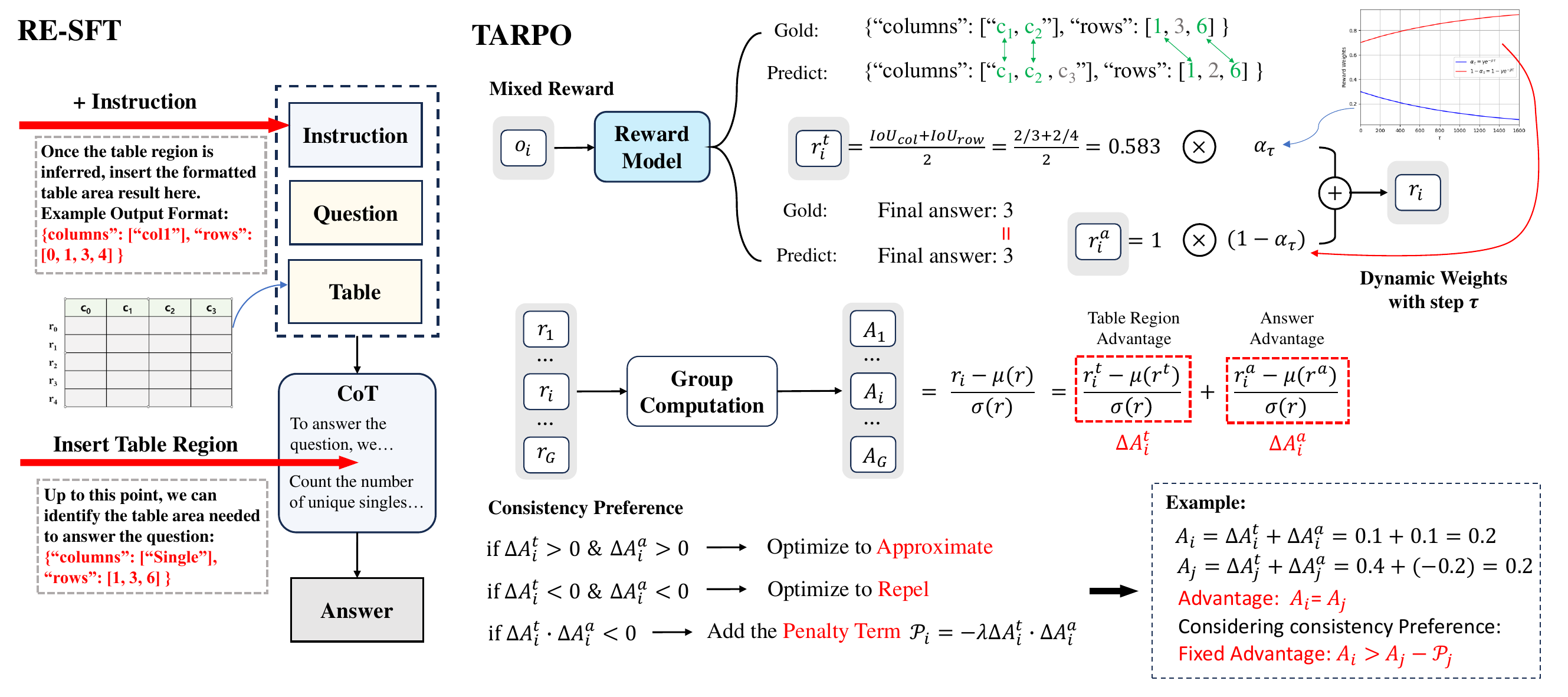}}
\caption{The framework of \ourmethod{}. In RE-SFT, we integrate the minimum table region at the correct position in the CoT process and add relevant instructions in the LLM prompt. In TARPO, we introduce mixed reward to balance the table region result and the answer result, and use $\alpha_\tau$ to dynamically adjust the weights in training. }
\label{overall}
\end{figure*}

\subsection{Problem Definition}
Table question answering (Table QA) is defined as: Given a semi-structured table $\mathcal{T}$ with $R$ rows and $C$ columns, the objective is to generate an answer $\mathcal{A}$ to a question $\mathcal{Q}$ by leveraging the information within $\mathcal{T}$. Here, $\mathcal{A}$ is a set of values or entities expressed as $\{a_1, a_2, \dots, a_k\}$ ($k \in \mathbb{N}^+$).

\subsection{Table Region Definition}
To concisely and intuitively describe a structured Table Region, we define it as: $T_{reg} = \{[c_1,c_2,...,c_i],[r_1,r_2,..,r_j]\}$. Figure~\ref{table region} illustrates why we use col \& row-based structure and present an example of how the LLM extracts relevant table region. Compared to the cell-based structure, col \& row-based structure not only saves input tokens but also preserves the sub-table structure. Though it may inevitably include some irrelevant cells, it still significantly narrows the data-acquisition scope for the LLM when computing answers. In the specific example, by combining the column attributes and cell values of the table with the target question, LLM can select the minimal table region required to generate the final answer. This approach efficiently narrows down the relevant data, focusing on the "Single" column and the corresponding rows that match the query criteria.
   
\subsection{Region-Enhanced Supervised Fine-Tuning}
In this section, we present \textbf{RE-SFT}, a Region-Enhanced Supervised Fine-Tuning approach. It aims to steer LLMs toward first identifying the table region relevant to answering questions during the CoT process. Subsequently, by leveraging the data from these identified table regions, the LLMs can compute and arrive at the final answers. We formalize the ICL (In-context Learning) process as $\mathcal{M}(\mathcal{I}(\mathcal{T}, \mathcal{Q}),\mathcal{E})$, where $\mathcal{I}$ denotes the task-specific instruction, $\mathcal{E}$ represents a few output examples, and $\mathcal{M}$ refers to the LLM. 

Tablebench~\citep{tablebench} has summarized four reasoning methods in TableQA: Direct Prompting (DP), Textual Chain-of-Thought (TCoT), Symbolic Chain-of-Thought (SCoT), and Program-of-Thoughts (PoT). We integrate the minimal table region $T_{reg}$ into these four methods:


{\small
\begin{flalign}
DP: \mathcal{M}(\mathcal{I}_{D}(\mathcal{T}, \mathcal{Q}),\mathcal{E}) \rightarrow \{T_{reg},\mathcal{A}\} \qquad \qquad \quad
\end{flalign}

\begin{equation}
\begin{split}
TCoT: \mathcal{M}&(\mathcal{I}_{T}(\mathcal{T}, \mathcal{Q}),\mathcal{E}) \rightarrow \\ \{&r_1, r_2, \ldots,r_m,T_{reg}, r_{m+1},\ldots, r_k, \mathcal{A}\}
\end{split}
\end{equation}

\begin{equation}
\begin{split}
S&CoT: \mathcal{M}(\mathcal{I}_{S}(\mathcal{T}, \mathcal{Q}),\mathcal{E}) \rightarrow \\& \{(r_{a_1}, r_{p_1}, r_{s_1}), \ldots, T_{reg}, \ldots, (r_{a_k}, r_{p_k}, r_{s_k}), \mathcal{A}\}
\end{split}
\end{equation}

\begin{flalign}
PoT: \mathcal{M}(\mathcal{I}_{P}(\mathcal{T}, \mathcal{Q}),\mathcal{E}) \rightarrow \{T_{reg},\mathcal{P}\} \rightarrow \mathcal{A} \qquad
\end{flalign}}
where $r_k$ is the $k$-th reasoning step in TCoT. $r_{a_k}$ is the analyzing step, $r_{p_k}$ is the program commands generating step, and $r_{s_k}$ is the result simulation step in SCoT. $\mathcal{P}$ is the generated code in PoT.

The training objective $\mathcal{L}_{\text{all}}$ of RE-SFT on datasets containing all four reasoning methods:

{\small
\begin{equation}
\mathcal{L}_{\text{all}} = -\sum_{n=1}^{N} \mathbb{E}_{q^{R_n}, a^{R_n}} \left[ \log P(a^{R_n} | q^{R_n}; \mathcal{M}) \right]
\end{equation}}

where \( q^{R_n} \) and \( a^{R_n} \) denote the table-related question and answer from dataset \( D^{R_n} \) of reasoning method \( R_n \), respectively. \( N \) represents the total number of reasoning methods.

\subsection{Table-Aware Reinforcement Learning}
Recently, reinforcement learning has shown remarkable performance on numerous tasks. In the specific tableQA task, we intend to offer additional rewards to the reasoning process that accurately identifies the relevant table region for better answer generation. Therefore, we have extended GRPO to develop the \textbf{TARPO} (Table-Aware Group Relative Policy Optimization) algorithm, which provides joint incentives for both the correctness of table region identification and final answer generation.

\subsubsection{Mixed Reward}
The reward $r_i$ for reasoning process $i$ is obtained by weighting the sum of the table region reward $r_i^{t}$ and the answer reward $r_i^{a}$:

\begin{equation}
r_i = \alpha_\tau  r_i^{t} + (1-\alpha_\tau)  r_i^{a}
\end{equation}

where $r_i^{a}$ is binary, taking a value of 1 when the inferred answer matches the correct answer, and 0 otherwise (Numerical answers match if equal, while string answers match if the Rouge-L score exceeds threshold $\zeta$). $r_i^{t}$ is calculated by computing the  IoU (Intersection over Union)~\citep{DBLP:conf/cvpr/RedmonF17} separately for the rows and columns between the table region obtained in the reasoning process and the ground truth. The results are then summed and averaged, with $r_i^{t}\in [0,1]$ :

{\small
\begin{equation}
r_i^{t} = \frac{IoU_{col}(T_{reg}, \hat{T_{reg}}) + IoU_{row}(T_{reg}, \hat{T_{reg}})}{2}
\end{equation}
}
where $IOU_{col}$ and $IOU_{row}$ respectively denote the IOU functions specific to rows and columns. $\hat{T_{reg}}$ stands for the predicted table region, while $T_{reg}$ represents the ground truth of the table region.

\subsubsection{Dynamic Weight of Reward}
At the beginning of reinforcement learning, our goal is to capture table regions more accurately to assist with answer reasoning. However, as training progresses and the model acquires a certain level of table region selection capability, we shift the focus of the reward more towards the accuracy of the answers. Thus, we have designed a dynamically changing weight $\alpha_\tau$ , which decreases with increasing training step $\tau$, the formula is as follows:

\begin{equation}
\alpha_\tau = \gamma e^{-\rho \tau}
\end{equation}

where $\gamma \in [0,1]$ is the initial weight, $\rho > 0$ is the decay coefficient.

\subsubsection{Consistency Preference}
Following GRPO, we can obtain the normalized group reward score ${A_i}$ by calculating the variance and standard deviation of the rewards. Combining equation (6), we can further decompose ${A}_i$ into two components $\Delta A_i^t$ and $\Delta A_i^a$:

{\small\begin{equation}
\begin{split}
A_i =& \frac{r_i - \text{mean}(\{r_1, r_2, \ldots, r_G\})}{\text{std}(\{r_1, r_2, \ldots, r_G\})}\\=&\underbrace{\frac{\alpha_\tau[r_i^{t} - \text{mean}(\{r_1^{t}, r_2^{t}, \ldots, r_G^{t}\})]}{\text{std}(\{r_1, r_2, \ldots, r_G\})}}_{\Delta A_i^t}+\\&\underbrace{\frac{(1-\alpha_\tau)[r_i^{a} - \text{mean}(\{r_1^{a}, r_2^{a}, \ldots, r_G^{a}\})]}{\text{std}(\{r_1, r_2, \ldots, r_G\})}}_{\Delta A_i^a}
\end{split}
\end{equation}}

Since the correct table region can guide better answer generation, we prefer the model to optimize in the direction where both $\Delta A_i^t$ and $\Delta A_i^a$ are positive. If $\Delta A_i^t$ and $\Delta A_i^a$ are of opposite signs, we consider that the reasoning process might be accidental and violates the consistency preference. Therefore, we introduce a small penalty $\mathcal{P}_i$ to mitigate the tendency of optimizing towards strategies (or accelerate the optimization away from it) that do not satisfy consistency:

{\small
\begin{equation}
\mathcal{P}_i = 
\begin{cases} 
0 &  \text{if } \Delta A_i^t \cdot \Delta A_i^a >0\\
-\lambda\Delta A_i^t \cdot \Delta A_i^a & \text{if } \Delta A_i^t \cdot \Delta A_i^a < 0
\end{cases}
\end{equation}
}

Incorporating the penalty term $\mathcal{P}_i$, the objective of TARPO is:

{\small
\begin{equation}
\begin{split}
J_{\text{TARPO}}&(\theta) = \mathbb{E}_{q, \{o_i\}_{i=1}^G} \Bigg[ \frac{1}{G} \sum_{i=1}^G \min \Bigg( \frac{\pi_{\theta}(o_i|q)}{\pi_{\theta_{\text{old}}}(o_i|q)} \\&(A_i - \mathcal{P}_i), 
\text{clip} \left( \frac{\pi_{\theta}(o_i|q)}{\pi_{\theta_{\text{old}}}(o_i|q)}, 1 - \epsilon, 1 + \epsilon \right) \\&(A_i - \mathcal{P}_i) \Bigg) - \beta D_{\text{KL}}(\pi_{\theta} \|\pi_{\text{ref}}) \Bigg]
\end{split}
\end{equation}
}

\section{Experimental Setup}
\label{sec:settings}
\textbf{Datasets.} We use \textbf{TableInstruct}~\citep{tablebench} as the training set because it encompasses all four reasoning methods, includes four major and 18 sub-question categories with the total size of nearly 20,000. For testing purposes, we employed three benchmark datasets. \textbf{TableBench}~\citep{tablebench} is a comprehensive tabular benchmark designed to evaluate large language models across multiple table-related tasks. \textbf{WikiTQ}~\citep{wikitq} is a dataset for table question answering, featuring 22,033 question-answer pairs with 2,108 tables. \textbf{WikiSQL}~\citep{zhong2017seq2sql} is a dataset that annotates Wiki tables with SQL, which contains 81,000 questions and 24,000 tables. The dataset statistics are summarized in Table~\ref{datasets}.

\begin{table}[tp]
\begin{center}
\resizebox{0.45 \textwidth}{!}{
\begin{tabular}{cccccc}
\bottomrule
\hline
\specialrule{0em}{1pt}{0pt}\raisebox{-2pt}[0pt][0pt]{
\multirow{2}{*} {Dataset}} & \raisebox{-2pt}[0pt][0pt]{\multirow{2}{*} {Tables}} & \raisebox{-2pt}[0pt][0pt]{\multirow{2}{*} {Samples}} & \multicolumn{3}{c}{Input Token Length}  \\
\specialrule{0em}{1pt}{0pt}
\cline {4-6} \specialrule{0em}{1pt}{0pt} & & & \quad min  \quad & max & median \\
\midrule
\specialrule{0em}{1pt}{0pt}
TableInstruct & 1.1K & 19,661 & 242 & 7,539 & 755\\
\hline
\specialrule{0em}{1pt}{0pt}
TableBench & 0.59K & 3,544 & 324 & 8,192 & 776\\
WikiTQ (test) & 0.4K & 4,344 & 266 & 2,120 & 691\\
WikiSQL (test) & 5K & 15,878 & 204 & 2,034 & 566\\

\bottomrule
\hline
\end{tabular}}
\end{center}
\caption{Statistics of datasets.}
\label{datasets}
\end{table}

\textbf{Evaluation Metrics.} We utilize the official evaluation metrics and codes of each benchmark dataset. For TableBench, we apply a combination of \textbf{Exact Match (EM)}, \textbf{Rouge-L}~\citep{rouge} and \textbf{Pass@1}~\citep{pass@1} metrics, while for WikiTQ and WikiSQL, we evaluate \textbf{Accuracy}.

\textbf{Baselines.} We select several competitive LLMs similar in size as baselines. The first category is composed of the most advanced open-source general-purpose LLMs, including DeepSeek-Coder-V2-Lite-16B~\citep{deepseek-v2}, Yi Coder-9B-Chat~\citep{yicoder}, Qwen2.5-Coder-7B-Instruct~\citep{qwen25coder}, Qwen2.5-7B-Instruct~\citep{qwen25} and QWQ-32B~\citep{qwen3}. We also use GPT-4o~\citep{gpt4o} as a powerful baseline. The second category comprises LLMs specifically designed for table-related tasks. TableLLM-13B~\citep{tablellm} fine-tuned CodeLlama-13B~\citep{codellama} to handle a variety of table operations in spreadsheet and document settings. CHAIN-OF-TABLE~\citep{DBLP:conf/iclr/0002ZLEP0MFSLP24} represents a tabular reasoning chain through in-context learning. TableGPT2-7B~\citep{tablegpt2} specifically designed a novel table encoder to capture schema-level and cell-level information. To better illustrate the improvement of \ourmethod{}, we introduce a third category of same-base LLMs that use the identical TableInstruct as the training set, including TableLLMs~\citep{tablebench} fine-tuned on Qwen2-7B~\citep{qwen2}, CodeQwen-7B~\citep{qwen}, Deepseek-Coder-7B~\citep{deepseek_coder}, Llama3.1-8B~\citep{llama3}, TeleChat2~\citep{telechat2}, Qwen2.5-3B~\citep{qwen25}, and Qwen3-8B~\citep{qwen3}.

\begin{table*}[ht]
\begin{center}
\resizebox{0.92 \textwidth}{!}{
\begin{tabular}{lccccccccc}
\bottomrule
\hline
\specialrule{0em}{1pt}{0pt}
\raisebox{-2pt}[0pt][0pt]{\multirow{2}{*} {Model}}& 
\raisebox{-2pt}[0pt][0pt]{\multirow{2}{*} {Base Model}}& \raisebox{-2pt}[0pt][0pt]{\multirow{2}{*} {Size}} & \multicolumn{4}{c}{TableBench} & \raisebox{-2pt}[0pt][0pt]{\multirow{2}{*} {WikiTQ}} & \raisebox{-2pt}[0pt][0pt]{\multirow{2}{*} {WikiSQL}}& \raisebox{-2pt}[0pt][0pt]{\multirow{2}{*} {Overall}}\\
\cmidrule(r){4-7}
&&& DP & TCoT & SCoT & PoT \\
\hline
\specialrule{0em}{1pt}{0pt}
\rowcolor{gray!15}
\multicolumn{10}{c} {General-purpose LLMs} \\
\hline
\specialrule{0em}{1pt}{0pt}
Yi-Coder & Yi & 9B & 21.94 & 22.80 & 8.43 & 11.36 & 43.37 & 25.34&22.21\\
DS-lite & DS-Coder & 16B & 29.60 & 30.93 & 22.61 & 10.90 & 47.65 & 38.30&30.00\\
Qwen2.5-Instruct & Qwen2.5 & 7B & 25.18 & 29.77 & 24.35 & {22.58} & {68.55} & {47.42}&36.14 \\
Qwen2.5-Coder & Qwen2.5 & 7B & {28.67} & {36.25} & {25.95} & 16.15 & {74.50} & 46.90&38.07\\
QWQ & Qwen2.5 & 32B & 43.87 &43.48& 37.06& 31.58 & 70.50 & 47.67& 45.69\\
GPT-4o & - & - & 40.91&51.96&41.43&45.71 & 68.40 & 47.60&49.17\\
\hline
\specialrule{0em}{1pt}{0pt}
\rowcolor{gray!15}
\multicolumn{10}{c} {Table-focused LLMs} \\
\hline
\specialrule{0em}{1pt}{0pt}
TableLLM & CodeLlama & 13B &  3.88 & 3.85 &  2.88 & 2.94 & 66.30 & 41.10&20.16 \\
CHAIN-OF-TABLE & GPT3.5 & - &  {24.86} & {32.61} & {26.39} & {20.24} & 59.94 & 43.72 & 34.63 \\
TableGPT2 & Qwen2.5 & 7B &  {27.95} & {41.05} & {31.4} & {38.67} & 61.42 & {53.74}&42.20 \\
TableGPT2 & Qwen2.5 & 72B &  {38.90} & {50.06} & {30.47} & {28.98} & 71.45 & {57.32}&46.20 \\
\hline
\specialrule{0em}{1pt}{0pt}
\rowcolor{gray!15}
\multicolumn{10}{c} {The same-base LLMs training on TableInstruct} \\
\hline
\specialrule{0em}{1pt}{0pt}
TableLLM (w/ SFT) & \multirow{3}{*} {Qwen2} & \multirow{3}{*} {7B} & 22.29 & 31.90 & 23.62 & 12.87 & \textbf{64.16} & 46.50&33.56\\
RE-TableLLM (w/ RE-SFT) &  & & \underline{28.24} & \underline{37.33} & \underline{29.10} & \underline{41.05} & 57.55 & \underline{64.42}&\underline{42.95}\\
\textbf{\ourmethod{}} (w/ RE-SFT \& TARPO) &  & &\textbf{36.22} & \textbf{41.85} & \textbf{32.84} & \textbf{41.13} & \underline{61.69} & \textbf{64.80}&\textbf{46.42}\\
\hline
\specialrule{0em}{1pt}{0pt}
TableLLM (w/ SFT) & \multirow{3}{*} {CodeQwen} & \multirow{3}{*} {7B} & 20.15&24.81&20.55&15.14&36.05&37.20&25.65\\
RE-TableLLM (w/ RE-SFT) &  & & \underline{21.47}&\underline{26.49}&\underline{24.34}&\textbf{42.52}&	\underline{46.71}&\underline{57.35}&\underline{36.48}\\
\textbf{\ourmethod{}} (w/ RE-SFT \& TARPO) &  & &\textbf{26.63}&\textbf{28.42}&\textbf{26.10}&\underline{41.15}&\textbf{49.31}&\textbf{59.89}&\textbf{38.58}\\
\hline
\specialrule{0em}{1pt}{0pt}
TableLLM (w/ SFT) & \multirow{3}{*} {DS-Coder} & \multirow{3}{*} {7B} & 23.15&30.51&23.56&18.74&	36.05&36.14&25.99\\
RE-TableLLM (w/ RE-SFT) &  & &\underline{25.63}&\underline{31.49}&\underline{25.88}&\underline{41.63}&\underline{47.88}&\underline{58.86}&\underline{36.13}\\
\textbf{\ourmethod{}} (w/ RE-SFT \& TARPO) &  & &\textbf{28.74}&\textbf{35.85}&\textbf{29.62}&\textbf{41.71}&\textbf{51.35}&\textbf{60.45}&\textbf{41.29}\\
\hline
\specialrule{0em}{1pt}{0pt}
TableLLM (w/ SFT) & \multirow{3}{*} {Llama3.1} & \multirow{3}{*} {8B} & 22.30&30.77&21.92&27.17&38.84&39.00&30.00\\
RE-TableLLM (w/ RE-SFT) &  & &\underline{28.71}&\underline{38.36}&\underline{30.46}&\underline{42.14}&\underline{61.03}&\underline{62.92}&\underline{43.94}\\
\textbf{\ourmethod{}} (w/ RE-SFT \& TARPO) &  & &\textbf{38.5}&\textbf{40.21}&\textbf{38.51}&\textbf{42.78}&\textbf{63.51}&\textbf{64.30}&\textbf{47.97}\\
\hline
\specialrule{0em}{1pt}{0pt}
TableLLM (w/ SFT) & \multirow{3}{*} {TeleChat2} & \multirow{3}{*} {7B} & 22.45&32.65&23.91&25.29&63.37&\underline{65.58}&38.88\\
RE-TableLLM (w/ RE-SFT) &  & &\underline{29.01}&\underline{38.57}&\underline{31.15}&\textbf{41.86}&\underline{63.52}&{64.93}&\underline{44.84}\\
\textbf{\ourmethod{}} (w/ RE-SFT \& TARPO) &  & &\textbf{39.29}&\textbf{42.14}&\textbf{38.95}&\underline{41.60}&\textbf{65.27}&\textbf{66.12}&\textbf{48.90}\\
\hline
\specialrule{0em}{1pt}{0pt}
TableLLM (w/ SFT) & \multirow{3}{*} {Qwen2.5} & \multirow{3}{*} {3B} & 20.08&30.02&22.11&27.55&\underline{55.50}&61.46&36.12\\
RE-TableLLM (w/ RE-SFT) &  & &\underline{22.38}&\underline{32.60}&\underline{25.57}&\underline{41.64}&{51.57}&\underline{61.55}&\underline{39.22}\\
\textbf{\ourmethod{}} (w/ RE-SFT \& TARPO) &  & &\textbf{35.71}&\textbf{39.54}&\textbf{36.78}&\textbf{41.93}&\textbf{56.51}&\textbf{63.98}&\textbf{45.74}\\
\hline
\specialrule{0em}{1pt}{0pt}
TableLLM (w/ SFT) & \multirow{3}{*} {Qwen3} & \multirow{3}{*} {8B} & 22.42&35.88&27.58&29.15&\underline{73.02}&\underline{71.14}&43.20\\
RE-TableLLM (w/ RE-SFT) &  & &\underline{31.53}&\underline{44.67}&\underline{34.70}&\underline{44.36}&69.06&69.57&\underline{48.98}\\
\textbf{\ourmethod{}} (w/ RE-SFT \& TARPO) &  & &\textbf{44.54}&\textbf{49.30}&\textbf{47.21}&\textbf{44.54}&\textbf{73.87}&\textbf{72.50}&\textbf{55.33}\\

\bottomrule
\hline
\end{tabular}}
\end{center}
\caption{The experimental results on three datasets. We compare \ourmethod{} LLMs with general-purpose LLMs, table-focused LLMs, and same-base LLMs. TableLLM-RE indicates the model built on TableLLM with only RE-SFT training, while \ourmethod{} represents our full framework, including both RE-SFT and TARPO. The experimental results demonstrate the effectiveness of RE-SFT and TARPO and and shows that \ourmethod{} achieves better performance with different base models. Bold numbers indicate the best result within each experimental group.}
\label{exp}

\end{table*}

\textbf{Implementation Details.} \textbf{(1) For Datasets:} We employ DeepSeek-R1 to generate the minimal table regions that suffice for each critical reasoning  step in the original TableInstruct dataset, manually verify the correctness, and obtain the  \textbf{TableInstruct-RE} dataset. We also add instructions related to table region generation in the prompts of both the training and test sets. As WikiSQL and WikiTQ lack ICL prompts, we use the modified TCoT prompt template of TableBench for them. \textbf{(2) For RE-SFT:} We achieve supervised fine-tuning on TableInstruct-RE dataset. Following the same setting of TableBench\citep{tablebench}, we use the entire dataset for training purposes and do not partition a validation set. We utilize a cosine annealing scheduler, which sets the initial learning rate at $2e^{-5}$, and conduct training over 3 epochs. Optimization is performed using the Adam optimizer, with a batch size of 512 and a maximum sequence length of 4096. We use 8*A100 GPUs for training, which takes 2 to 2.5 hours for each training. \textbf{(3) For TARPO:}  We randomly shuffled the TableInstruct-RE dataset and split it into training and validation sets in a ratio of 9:1. We set the batch size to 32, with the max prompt length and max response length at 8192 and 2048. The learning rate is $7e^{-7}$, and the group number  $G$  for each question is 16. The 7B/8B-sized LLMs are trained on 4*80GB-H100 GPUs, and the 3B-sized LLM on 8*40GB-A100 GPUs. The training lasts for 3 epochs and around 72 hours. For the hyper parameters, we set \{$\zeta$, $\gamma$, $\rho$, $\lambda$\} to \{0.6, 
 0.3, $9e^{-4}$, 0.1\}.

\begin{table*}[ht]
\begin{center}
\resizebox{0.95 \textwidth}{!}{
\begin{tabular}{lccccccccc}
\bottomrule
\hline
\specialrule{0em}{1pt}{0pt}
\raisebox{-2pt}[0pt][0pt]{\multirow{2}{*} {Model}} & \multicolumn{5}{c}{TableBench} & \multicolumn{2}{c}{WikiTQ} & \multicolumn{2}{c}{WikiSQL}\\
\cmidrule(r){2-6}\cmidrule(r){7-8}\cmidrule(r){9-10}
& DP & TCoT & SCoT & PoT & Avg tokens & Acc & Avg tokens & Acc & Avg tokens \\
\hline
\specialrule{0em}{2pt}{0pt}
w/ SFT (TableLLM) &22.42&35.88&27.58&29.15&\textbf{581}&73.02&\textbf{313}&71.14&\textbf{201}\\
\hline
\specialrule{0em}{2pt}{0pt}
\multirow{2}{*} {w/ RE-SFT} &31.53&44.67&34.70&44.36&1,330&69.06&1,028&69.57&296\\
&\textcolor{blue}{($\uparrow$ 9.81)}&\textcolor{blue}{($\uparrow$ 8.79)}&\textcolor{blue}{($\uparrow$ 7.12)}&\textcolor{blue}{($\uparrow$ 15.21)}&\textcolor{red}{($\uparrow$ 129\%)}&{($\downarrow$ 3.96)}&\textcolor{red}{($\uparrow$ 228\%)}&{($\downarrow$ 1.57)}&\textcolor{red}{($\uparrow$ 47.3\%)}\\
\hline
\specialrule{0em}{2pt}{0pt}
\multirow{2}{*} {w/ RE-SFT \& GRPO}&42.41&48.64&45.62&42.44&1,549&\underline{73.15}&996&70.24&1,399\\
&\textcolor{blue}{($\uparrow$ 10.88)}&\textcolor{blue}{($\uparrow$ 3.97)}&\textcolor{blue}{($\uparrow$ 10.92)}&{($\downarrow$ 1.92)}&\textcolor{red}{($\uparrow$ 16.5\%)}&\textcolor{blue}{($\uparrow$ 4.09)}&{($\downarrow$ 0.03\%)}&{($\uparrow$ 0.67)}&\textcolor{red}{($\uparrow$ 373\%)}\\
\hline
\specialrule{0em}{2pt}{0pt}
\multirow{2}{*} {w/ RE-SFT \& TARPO (w/o $\mathcal{P}$)}&\underline{44.43}&\underline{49.15}&\underline{46.88}&\underline{44.20}&630&73.09&416&\underline{70.83}&276\\
&{($\uparrow$ 2.02)}&{($\uparrow$ 0.51)}&{($\uparrow$ 1.26)}&{($\uparrow$ 1.76)}&\textcolor{blue}{($\downarrow$ 59.3\%)}&{($\downarrow$ 0.06)}&\textcolor{blue}{($\downarrow$ 58.2\%)}&{($\uparrow$ 0.59)}&\textcolor{blue}{($\downarrow$ 80.3\%)}\\
\hline
\specialrule{0em}{2pt}{0pt}
\multirow{2}{*} {w/ RE-SFT \& TARPO (w/ $\mathcal{P}$)}&\textbf{44.54}&\textbf{49.30}&\textbf{47.21}&\textbf{44.54}&\underline{612}&\textbf{73.87}&\underline{397}&\textbf{72.50}&\underline{251}\\
&{($\uparrow$ 0.11)}&{($\uparrow$ 0.15)}&{($\uparrow$ 0.33)}&{($\uparrow$ 0.34)}&{($\downarrow$ 0.03\%)}&\textcolor{blue}{($\uparrow$ 0.78)}&{($\downarrow$ 0.05\%)}&\textcolor{blue}{($\uparrow$ 1.67)}&{($\downarrow$ 0.09\%)}\\

\bottomrule
\hline
\end{tabular}}
\end{center}
\caption{The ablation study of \ourmethod{} based on Qwen3-8B. Blue indicates better performance (higher scores and lower average tokens), while red shows worse performance (lower scores and higher average tokens).}
\label{exp2}
\end{table*}

\section{Results and Analysis}
\label{sec:results}

\subsection{Overall Results}
Table~\ref{exp} shows the TableBench, WikiTQ, and WikiSQL results for baselines and our \ourmethod{}.
We calculate an overall score from the average test-set scores to evaluate LLM performance. Among general-purpose LLMs, the small-sized (under 16B) Qwen2.5 series shows strong overall performance, with Qwen2.5-Coder scoring 38.07. Specifically, the Qwen2.5-Coder achieves the overall score of 38.07. Larger models like GPT-4o achieve a higher overall score of 49.17 and perform well across various datasets. For table-focused LLMs, TableGPT2-7B model stands out in small models with a 42.2 overall score, surpassing general-purpose LLMs and TableLLMs of the same size. The larger TableGPT2-72B further boosts the overall score to 46.2.

We implement the \ourmethod{} method across different LLM architectures and compare its performance with TableLLMs trained on the same TableInstruct training dataset. Each experimental group consisted of three model configurations: 1) the baseline TableLLM, 2) RE-SFT introduced via supervised fine-tuning, and 3) the model with both RE-SFT and TARPO (\ourmethod{}). This design systematically evaluates RE-SFT and TARPO’s contributions. Results show RE-SFT delivers overall performance gains across three benchmarks, especially in TableBench PoT data, with minor declines observed on cross-domain test sets (WikiTQ and WikiSQL). Across seven models, RE-SFT achieves an average PoT score increase of 19.90 and an overall increase of 8.45. TARPO integration further enhances performance.  As shown in Table~\ref{exp}, \ourmethod{} achieves significant improvement on DP, TCoT, and SCoT data of TableBench, WikiTQ and WikiSQL datasets, achieving an average gain of 4.52 in overall score than RE-SFT alone.
In conclusion, \ourmethod{} gets the top-tier performance on different benchmarks by a large margin. Especially, with the same base model Qwen2.5, 3B-sized \ourmethod{} surpasses 7B-sized Qwen2.5-Coder, Qwen2.5-Instruct and TableGPT2 overall. On Qwen3-8B, \ourmethod{} exceeds GPT-4o by 6.16 overall, underperforming only by 0.76 on TCoT.

\subsection{Ablation Study}
We use Qwen3-8B as the base model for ablation study. Table~\ref{exp2} presents the experimental results.

Compared to the baseline TableLLM~\citep{tablebench}, RE-SFT significantly enhances the model's performance on TableBench, especially with a 15.21 gain in PoT data. Yet, it causes performance drops of 3.96 and 1.57 on the other two out-of-domain test sets WikiTQ and WikiSQL. Integrating table regions into CoT also increases average response tokens across the three test sets by 129\%, 228\%, and 47.3\% respectively. We further employ GRPO algorithm for reinforcement learning. Experiment results show marked improvement in Tablebench DP, TCoT, SCoT, and WikiTQ performance, while PoT performance slightly decreases. GRPO also causes a rise in response reasoning tokens, with TableBench and WikiSQL seeing average increases of 16.5\% and 373\%. Then, we substitute GRPO with the TARPO algorithm (without penalty $\mathcal{P}$). The results demonstrate slight performance enhancements on TableBench and WikiSQL, while experiencing a minor decline on WikiTQ. Notably, TARPO achieve a significant reduction in average response tokens compared to GRPO, with decreases of 59.3\%, 58.2\%, and 80.3\% across three datasets, which suggests that TARPO can guide a more concise and direct CoT process. Furthermore, we add the penalty $\mathcal{P}$ in TARPO to achieve consistency preference. The results show a slight improvement on TableBench and more noticeable enhancements on WikiTQ and WikiSQL by 0.78\% and 1.67\%. This indicates that the penalty $\mathcal{P}$ effectively constrain the optimization consistency between table region and answer in training. It strengthens the rationality of the reasoning process by preventing the LLM from learning low-quality reasoning strategies that generate correct answers accidentally. Consequently, it improves the generalization of LLM on out-of-domain datasets.

In conclusion, the whole \ourmethod{} framework achieves optimal performance across all three datasets. Meanwhile, when incorporating table regions, TARPO increases the average response tokens by 5.3\%, 26.8\%, and 24.9\% respectively on the three datasets compared to TableLLM, but reduces them by an average of 60.5\%, 60.1\%, and 82.1\% compared to the GRPO, which effectively addresses the excessive reasoning length caused by RE-SFT and GRPO.

\begin{figure*}[ht]
    \centering
    \begin{subfigure}{0.3\textwidth} 
        \centering
        \includegraphics[width=\linewidth]{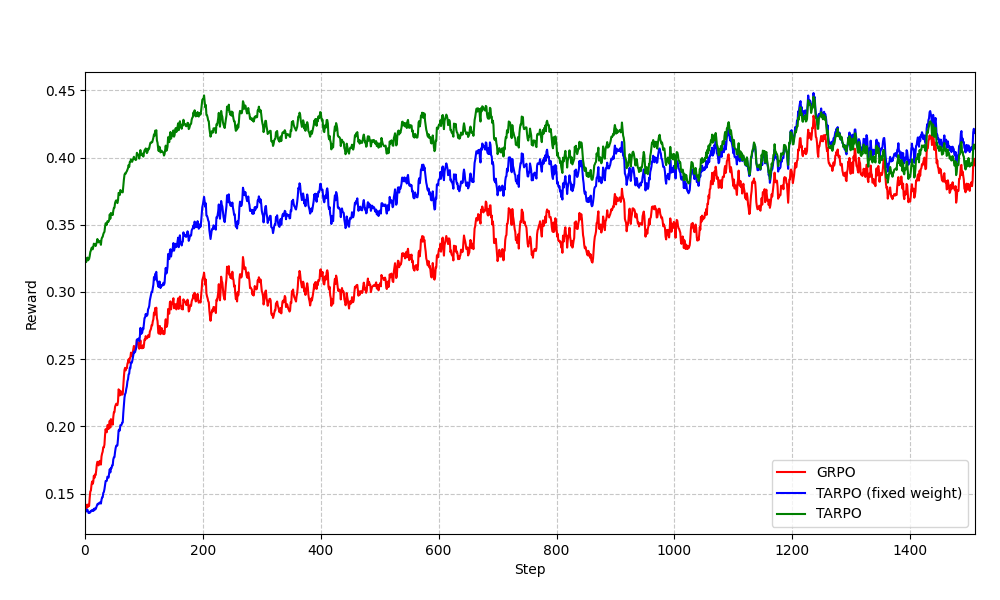}
        \caption{Mean Reward}
        \label{fig:figure1}
    \end{subfigure}
    \hfill 
    \begin{subfigure}{0.3\textwidth}
        \centering
        \includegraphics[width=\linewidth]{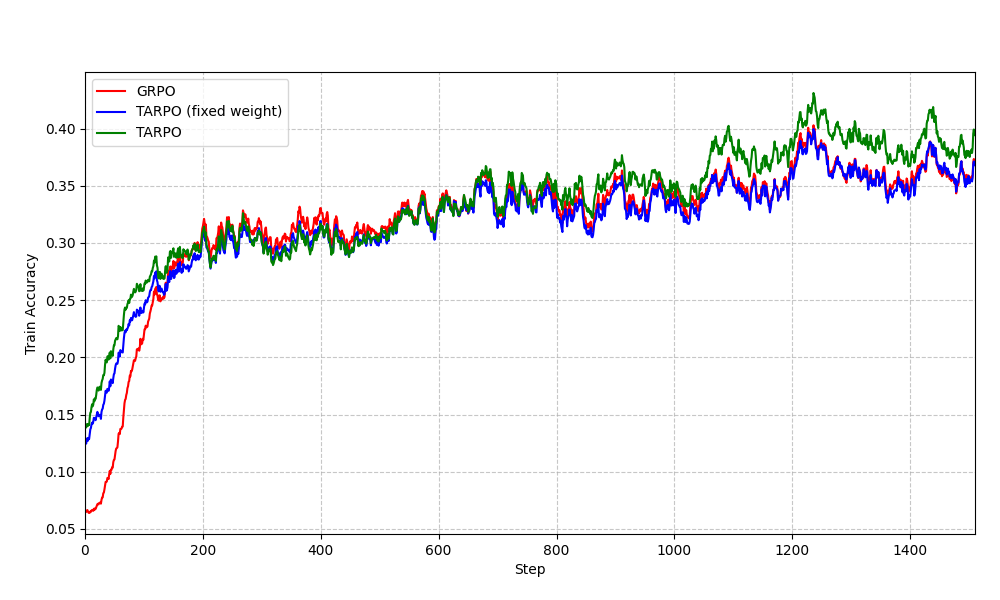}
        \caption{Train Accuracy}
        \label{fig:figure2}
    \end{subfigure}
    \hfill 
    \begin{subfigure}{0.3\textwidth}
        \centering
        \includegraphics[width=\linewidth]{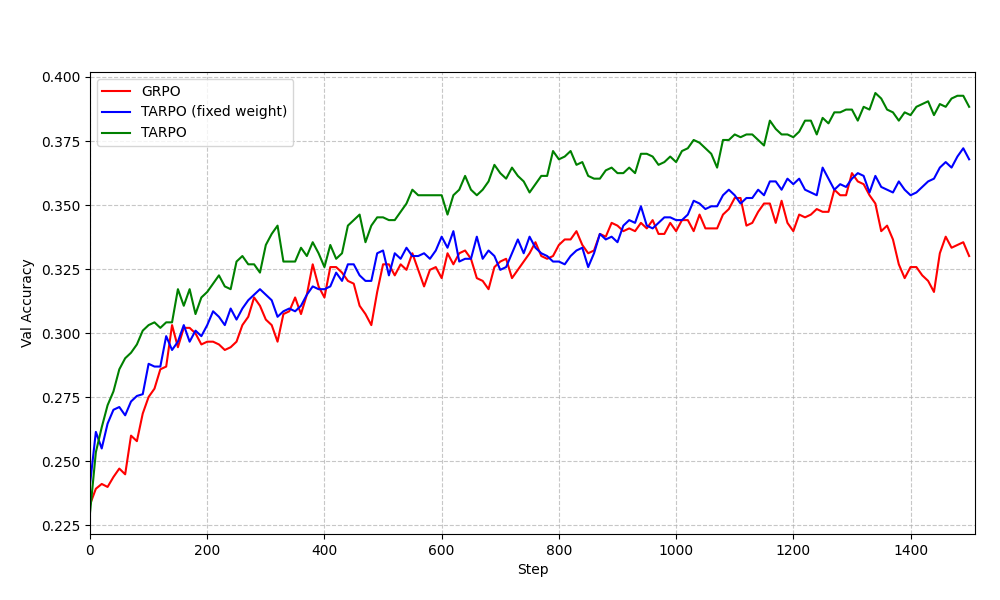}
        \caption{Val Accuracy}
        \label{fig:figure3}
    \end{subfigure}

    \vspace{1em} 
    \begin{subfigure}{0.3\textwidth} 
        \centering
        \includegraphics[width=\linewidth]{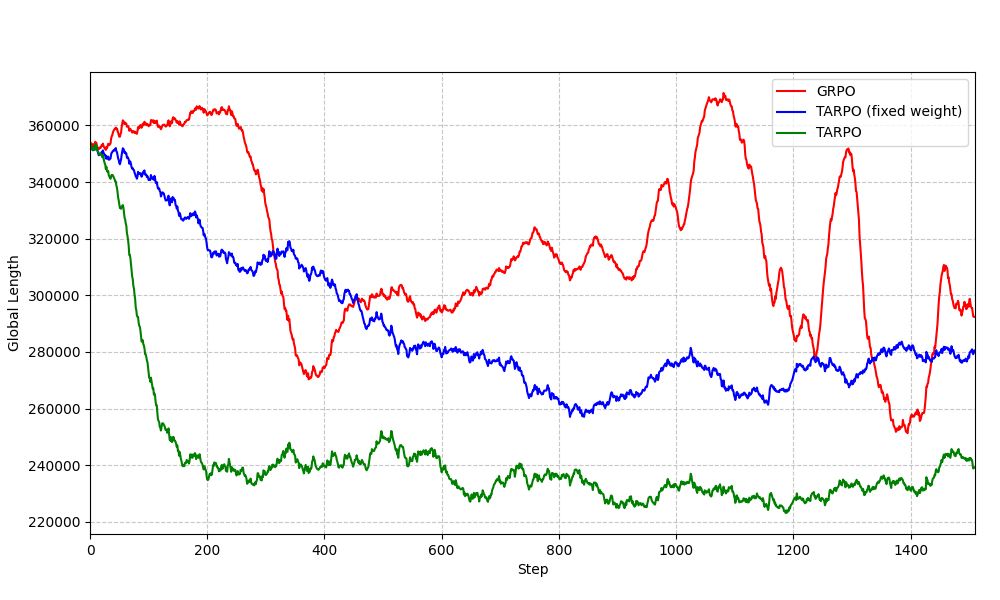}
        \caption{Global Length (bz=32)}
        \label{fig:figure4}
    \end{subfigure}
    \hfill 
    \begin{subfigure}{0.3\textwidth}
        \centering
        \includegraphics[width=\linewidth]{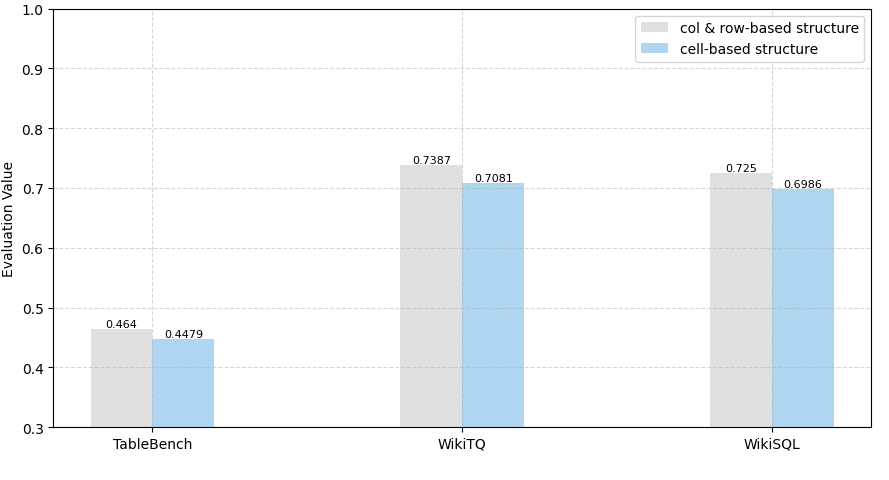}
        \caption{Evaluation Value}
        \label{fig:figure5}
    \end{subfigure}
    \hfill 
    \begin{subfigure}{0.3\textwidth}
        \centering
        \includegraphics[width=\linewidth]{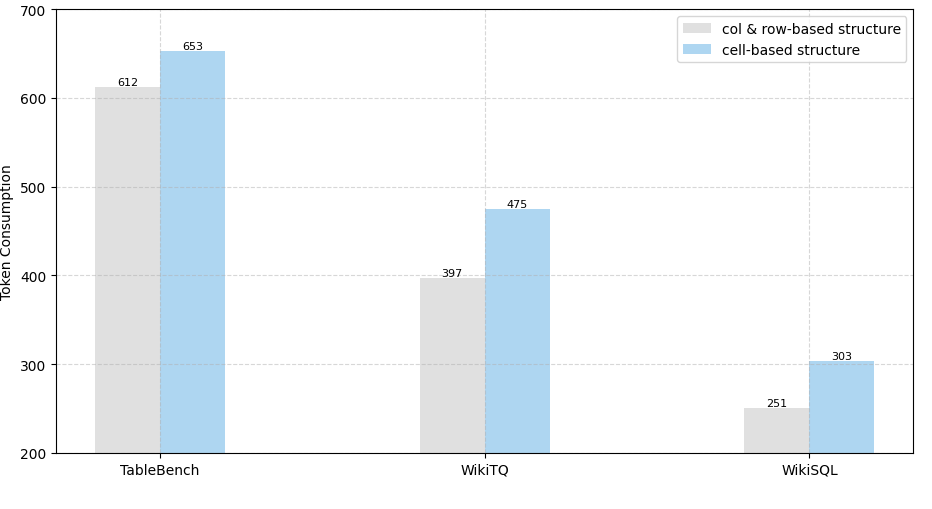}
        \caption{Token Consumption}
        \label{fig:figure6}
    \end{subfigure}

    \caption{(a)-(d) Reinforcement learning training data statistics for TableBench on Qwen3-8B. (e)-(f) The impact of col \& row-based table structure and cell-based structure on evaluation value and token consumption.}
    \label{stastics}
\end{figure*}

\subsection{Stastics Analysis}
\label{sec:further}
We compare the data stastics of GRPO, fixed-weight TARPO ($\alpha_\tau=0.15$), and TARPO during reinforcement training. Figure~\ref{stastics} shows the mean reward, train accuracy, valid accuracy, and global token length as step increases. To better observe the overall trends, we apply EMA (Exponential Moving Average) to the three metrics except validation accuracy, with a decay factor of 0.05. It can be seen that TARPO's high initial $\alpha_\tau$ value keeps its mean reward high in the early training phase. However, as $\alpha_\tau$ decreases with step, the final reward of TARPO ends up higher than GRPO but lower than fixed-weight TARPO. This is because fixed-weight TARPO continues to benefit from the reward of the table region part due to the retained $\alpha_\tau$ value at the end of training. Figure~\ref{stastics} (b) and (c) illustrate the accuracy of the current batch and the validation set during training, respectively. TARPO demonstrates better performance compared to the other two methods, indicating both its advantage over GRPO and the effectiveness of dynamic weights. Figure~\ref{stastics} (d) shows the global token length per step. TARPO shows a superior ability in reducing output token length compared to GRPO and also exhibits a more significant reduction than fixed-weight TARPO. Figure~\ref{stastics} (e) and (f) demonstrate that the col \& row-based table structure surpasses the cell-based structure in terms of performance and token conservation.

\section{Related Work}
\label{sec:relatedwork}
\textbf{Table Understanding.} The evolution of Table Question Answering (Table QA) research~\citep{mueller2019answering,alm,jin2022survey,qwen3,execrepobench,qwen25coder,DBLP:journals/tkde/LinRLW24,wu-feng-2024-protrix,MR-SQL,DBLP:conf/semweb/LiWLHFZZLLS24,UCS-SQL,ye2025tableqa,guo2026rethinking,Guo_2025,hong2025next,hong2024knowledge,yuan2025knapsack,zhang2024structure} has been propelled by the creation of sophisticated evaluation resources that facilitate semantic parsing capabilities~\citep{DBLP:conf/coling/HuangMLHZ0Z24,soft_template,DBLP:journals/corr/abs-2311-11268, DBLP:journals/corr/abs-2402-11100,DBLP:conf/acl/NakamuraLTCW22,cTBLS,qwen,qwen2,codearena,xcoder,xcot,DBLP:conf/naacl/ZhangYL024}.
Foundational works including WTQ~\citep{pasupat2015compositional}, SQA~\citep{iyyer2017search}, and TabFact~\citep{chen2019tabfact} established initial evaluation paradigms through Wikipedia-derived HTML table QA pairs. While these resources provide structured testing grounds, their answer formulations predominantly depend on direct cell retrieval, limiting their capacity to emulate the multifaceted information needs observed in practical applications.
Recent initiatives address this limitation through innovative dataset designs. ToTTo~\citep{parikh2020totto} and FeTaQA~\citep{nan2022fetaqa} pioneer open-domain QA frameworks requiring generative responses that synthesize implicit table relationships. Alternative approaches employ structured supervision signals, exemplified by WikiSQL~\citep{zhong2017seq2sql} and Spider~\citep{yu2018spider}, which utilize logical expressions to cultivate systematic reasoning skills.

\textbf{RL for LLMs.} Driven by the widespread expansion of LLM applications~\citep{chang2024survey,chang2026balora,dong2025aurora,dong2026neureason,jiang2026foefores,yang2026incoder,yang2026incoder_thinking,yang2025codesimpleqa,DBLP:conf/emnlp/YangYZKMZYCZLHL25,DBLP:conf/acl/YangZMQW0Y0CHL25,yang2025code,yang2026iquest}, recent advancements in reinforcement learning (RL) frameworks have demonstrated significant improvements in augmenting the inferential capacities of LLMs, exemplified by architectures like DeepSeek‑R1~\citep{deepseek_r1} and OpenAI‑o1~\citep{o1}. Driven by these breakthroughs, a burgeoning body of research has increasingly deployed reinforcement learning paradigms across diverse specialized domains, facilitating autonomous knowledge discovery and systematic exploration within complex task environments~\cite{wu2026step,zhao2026reinforced,zhou2026look,hao2025rethinking,liu2025uniform,TTVS,chang2025lora}. Through cyclical switching between exploratory response generation and strategic knowledge utilization, these frameworks progressively refine output granularity while achieving sustained performance gains. State-of-the-art implementations predominantly leverage policy optimization techniques, including Proximal Policy Optimization (PPO)~\citep{ppo} and the parameter-efficient Group Relative Policy Optimization (GRPO)~\citep{grpo}. The latter methodology eliminates dependency on auxiliary value estimation networks, thereby achieving enhanced computational economy.

\section{Conclusion}
In this paper, we introduce \ourmethod{}, a novel region-based reinforcement learning method for table understanding. Through RE-SFT and TARPO training, \ourmethod{} effectively incorporates the extraction of the minimum table region into the reasoning CoT, thereby enhancing the guidance of answer generation. The experimental results show that \ourmethod{} delivers an average improvement of 14.36 points across multiple base models on three benchmarks. Moreover, TARPO reduces response token consumption by an average of 67.5\% compared to GRPO. In future work, we hope \ourmethod{} will inspire more research on rule-based reinforcement learning methods for different domain-specific tasks. We also aim to encourage the exploration of signals beyond final answers in the CoT process that can be used for reward shaping to enhance the quality of reasoning process.

\section*{Limitations}
Our experiments are conducted exclusively on models with 3B to 8B parameters and are limited to maximum input and output lengths of 8192 and 2048 tokens, respectively. The results may vary for larger models or longer token sequences. In addition, for simplicity, we only inserted the table region midway into CoT without refining the CoT process, the results may not be optimal. Our current study focuses on established academic datasets and has not yet been validated on real-world complex tables in industrial settings or multi-table QA scenarios.

\section*{Ethics Statement}
In this work, all of the datasets, models, code and related documents are not associated with any ethical concerns.

\section*{Acknowledgments}
This work was supported in part by the National Natural Science Foundation of China (Grant Nos. 62276017, 62406033, U1636211, 61672081), the State Key Laboratory of Complex \& Critical Software Environment (Grant Nos. SKLCCSE-2024ZX-18, SKLCCSE-2025ZX-26), and the Fundamental Research Funds for the Central Universities (Grant No. GW2025-19).


\bibliography{custom}

\begin{thebibliography}{88}
\providecommand{\natexlab}[1]{#1}

\bibitem[{01.AI(2024)}]{yicoder}
01.AI. 2024.
\newblock \href {https://github.com/01-ai/Yi-Coder} {Meet yi-coder: A small but mighty llm for code}.

\bibitem[{Anirudh S~Sundar(2023)}]{cTBLS}
Larry~Heck Anirudh S~Sundar. 2023.
\newblock ctbls: Augmenting large language models with conversational tables.
\newblock \emph{arXiv preprint arXiv:2303.12024}.

\bibitem[{Bai et~al.(2023)Bai, Bai, Chu, Cui, Dang, Deng, Fan, Ge, Han, Huang et~al.}]{qwen}
Jinze Bai, Shuai Bai, Yunfei Chu, Zeyu Cui, Kai Dang, Xiaodong Deng, Yang Fan, Wenbin Ge, Yu~Han, Fei Huang, and 1 others. 2023.
\newblock Qwen technical report.
\newblock \emph{arXiv preprint arXiv:2309.16609}.

\bibitem[{Bai et~al.(2026)Bai, Li, Zhang, Liu, and Guo}]{TTVS}
Sikai Bai, Haoxi Li, Jie Zhang, Yongjiang Liu, and Song Guo. 2026.
\newblock Ttvs: Boosting self-exploring reinforcement learning via test-time variational synthesis.

\bibitem[{Chai et~al.(2024)Chai, Yang, Sun, Guo, Liu, Wang, Liang, Bai, Li, Peng, and Li}]{xcot}
Linzheng Chai, Jian Yang, Tao Sun, Hongcheng Guo, Jiaheng Liu, Bing Wang, Xinnian Liang, Jiaqi Bai, Tongliang Li, Qiyao Peng, and Zhoujun Li. 2024.
\newblock \href {https://doi.org/10.48550/ARXIV.2401.07037} {xcot: Cross-lingual instruction tuning for cross-lingual chain-of-thought reasoning}.
\newblock \emph{arXiv preprint arXiv:2401.07037}, abs/2401.07037.

\bibitem[{Chang et~al.(2026)Chang, Chang, and Wu}]{chang2026balora}
Yupeng Chang, Yi~Chang, and Yuan Wu. 2026.
\newblock \href {https://openreview.net/forum?id=q0X9SiXiRO} {{BA}-lo{RA}: Bias-alleviating low-rank adaptation to mitigate catastrophic inheritance in large language models}.
\newblock In \emph{The Fourteenth International Conference on Learning Representations}.

\bibitem[{Chang et~al.(2025)Chang, Guo, Chang, and Wu}]{chang2025lora}
Yupeng Chang, Chenlu Guo, Yi~Chang, and Yuan Wu. 2025.
\newblock Lora-mgpo: Mitigating double descent in low-rank adaptation via momentum-guided perturbation optimization.
\newblock In \emph{Findings of the Association for Computational Linguistics: EMNLP 2025}, pages 648--659.

\bibitem[{Chang et~al.(2024)Chang, Wang, Wang, Wu, Yang, Zhu, Chen, Yi, Wang, Wang et~al.}]{chang2024survey}
Yupeng Chang, Xu~Wang, Jindong Wang, Yuan Wu, Linyi Yang, Kaijie Zhu, Hao Chen, Xiaoyuan Yi, Cunxiang Wang, Yidong Wang, and 1 others. 2024.
\newblock A survey on evaluation of large language models.
\newblock \emph{ACM transactions on intelligent systems and technology}, 15(3):1--45.

\bibitem[{Chen et~al.(2021)Chen, Tworek, Jun, Yuan, de~Oliveira~Pinto, Kaplan, Edwards, Burda, Joseph, Brockman et~al.}]{pass@1}
Mark Chen, Jerry Tworek, Heewoo Jun, Qiming Yuan, Henrique~Pond{\'{e}} de~Oliveira~Pinto, Jared Kaplan, Harri Edwards, Yuri Burda, Nicholas Joseph, Greg Brockman, and 1 others. 2021.
\newblock Evaluating large language models trained on code.
\newblock \emph{CoRR}, abs/2107.03374.

\bibitem[{Chen et~al.(2022)Chen, Ma, Wang, and Cohen}]{pot}
Wenhu Chen, Xueguang Ma, Xinyi Wang, and William~W. Cohen. 2022.
\newblock \href {https://doi.org/10.48550/ARXIV.2211.12588} {Program of thoughts prompting: Disentangling computation from reasoning for numerical reasoning tasks}.
\newblock \emph{arXiv preprint arXiv:2211.12588}, abs/2211.12588.

\bibitem[{Chen et~al.(2019)Chen, Wang, Chen, Zhang, Wang, Li, Zhou, and Wang}]{chen2019tabfact}
Wenhu Chen, Hongmin Wang, Jianshu Chen, Yunkai Zhang, Hong Wang, Shiyang Li, Xiyou Zhou, and William~Yang Wang. 2019.
\newblock Tabfact: A large-scale dataset for table-based fact verification.
\newblock \emph{CoRR}.

\bibitem[{Cheng et~al.(2023)Cheng, Xie, Shi, Li, Nadkarni, Hu, Xiong, Radev, Ostendorf, Zettlemoyer, Smith, and Yu}]{DBLP:conf/iclr/ChengX0LNHXROZS23}
Zhoujun Cheng, Tianbao Xie, Peng Shi, Chengzu Li, Rahul Nadkarni, Yushi Hu, Caiming Xiong, Dragomir Radev, Mari Ostendorf, Luke Zettlemoyer, Noah~A. Smith, and Tao Yu. 2023.
\newblock Binding language models in symbolic languages.
\newblock In \emph{The Eleventh International Conference on Learning Representations, {ICLR} 2023, Kigali, Rwanda, May 1-5, 2023}. OpenReview.net.

\bibitem[{DeepSeek{-}AI et~al.(2024)DeepSeek{-}AI, Liu, Feng, Wang, Wang, Liu, Zhao, Deng, Ruan, Dai, Guo et~al.}]{deepseek-v2}
DeepSeek{-}AI, Aixin Liu, Bei Feng, Bin Wang, Bingxuan Wang, Bo~Liu, Chenggang Zhao, Chengqi Deng, Chong Ruan, Damai Dai, Daya Guo, and 1 others. 2024.
\newblock Deepseek-v2: {A} strong, economical, and efficient mixture-of-experts language model.
\newblock \emph{CoRR}, abs/2405.04434.

\bibitem[{Dong et~al.(2026)Dong, Jiang, Ye, Zhu, Kang, and Song}]{dong2026neureason}
Haonan Dong, Kehan Jiang, Haoran Ye, Wenhao Zhu, Zhaolu Kang, and Guojie Song. 2026.
\newblock \href {https://arxiv.org/abs/2604.02972} {Neureasoner: Towards explainable, controllable, and unified reasoning via mixture-of-neurons}.
\newblock \emph{Preprint}, arXiv:2604.02972.

\bibitem[{Dong et~al.(2025)Dong, Zhu, Song, and Wang}]{dong2025aurora}
Haonan Dong, Wenhao Zhu, Guojie Song, and Liang Wang. 2025.
\newblock \href {https://openreview.net/forum?id=2hgHyoyVWj} {Auro{RA}: Breaking low-rank bottleneck of lo{RA} with nonlinear mapping}.
\newblock In \emph{The Thirty-ninth Annual Conference on Neural Information Processing Systems}.

\bibitem[{Dubey et~al.(2024)Dubey, Jauhri, Pandey, Kadian, Al-Dahle, Letman, Mathur, Schelten, Yang, Fan et~al.}]{llama3}
Abhimanyu Dubey, Abhinav Jauhri, Abhinav Pandey, Abhishek Kadian, Ahmad Al-Dahle, Aiesha Letman, Akhil Mathur, Alan Schelten, Amy Yang, Angela Fan, and 1 others. 2024.
\newblock The llama 3 herd of models.
\newblock \emph{arXiv preprint arXiv:2407.21783}.

\bibitem[{Guo et~al.(2025{\natexlab{a}})Guo, Yang, Zhang, Song, Zhang, Xu, Zhu, Ma, Wang, Bi et~al.}]{deepseek_r1}
Daya Guo, Dejian Yang, Haowei Zhang, Junxiao Song, Ruoyu Zhang, Runxin Xu, Qihao Zhu, Shirong Ma, Peiyi Wang, Xiao Bi, and 1 others. 2025{\natexlab{a}}.
\newblock Deepseek-r1: Incentivizing reasoning capability in llms via reinforcement learning.
\newblock \emph{arXiv preprint arXiv:2501.12948}.

\bibitem[{Guo et~al.(2024)Guo, Zhu, Yang, Xie, Dong, Zhang, Chen, Bi, Wu, Li et~al.}]{deepseek_coder}
Daya Guo, Qihao Zhu, Dejian Yang, Zhenda Xie, Kai Dong, Wentao Zhang, Guanting Chen, Xiao Bi, Y~Wu, YK~Li, and 1 others. 2024.
\newblock \href {https://arxiv.org/abs/2401.14196} {Deepseek-coder: When the large language model meets programming--the rise of code intelligence}.
\newblock \emph{arXiv preprint arXiv:2401.14196}.

\bibitem[{Guo et~al.(2025{\natexlab{b}})Guo, Jin, Ye, Chen, Jianyang, and Tan}]{Guo_2025}
Yu~Guo, Dong Jin, Shenghao Ye, Shuangwu Chen, Jianyang Jianyang, and Xiaobin Tan. 2025{\natexlab{b}}.
\newblock Sqlforge: Synthesizing reliable and diverse data to enhance text-to-sql reasoning in llms.
\newblock In \emph{Findings of the Association for Computational Linguistics: ACL 2025}.

\bibitem[{Guo et~al.(2026)Guo, Ye, Chen, Wen, Zhang, Bai, Jin, Hou, He, Yang et~al.}]{guo2026rethinking}
Yu~Guo, Shenghao Ye, Shuangwu Chen, Zijian Wen, Tao Zhang, Qirui Bai, Dong Jin, Yunpeng Hou, Huasen He, Jian Yang, and 1 others. 2026.
\newblock Rethinking table pruning in tableqa: From sequential revisions to gold trajectory-supervised parallel search.
\newblock \emph{arXiv preprint arXiv:2601.03851}.

\bibitem[{Hao et~al.(2025)Hao, Wang, Liu, Luo, Yu, Dong, Lin, Wang, and Chen}]{hao2025rethinking}
Zhezheng Hao, Hong Wang, Haoyang Liu, Jian Luo, Jiarui Yu, Hande Dong, Qiang Lin, Can Wang, and Jiawei Chen. 2025.
\newblock Rethinking entropy interventions in rlvr: An entropy change perspective.
\newblock \emph{arXiv preprint arXiv:2510.10150}.

\bibitem[{Hong et~al.(2024)Hong, Yuan, Chen, Zhang, Huang, and Huang}]{hong2024knowledge}
Zijin Hong, Zheng Yuan, Hao Chen, Qinggang Zhang, Feiran Huang, and Xiao Huang. 2024.
\newblock Knowledge-to-sql: Enhancing sql generation with data expert llm.
\newblock In \emph{Findings of the Association for Computational Linguistics: ACL 2024}, pages 10997--11008.

\bibitem[{Hong et~al.(2025)Hong, Yuan, Zhang, Chen, Dong, Huang, and Huang}]{hong2025next}
Zijin Hong, Zheng Yuan, Qinggang Zhang, Hao Chen, Junnan Dong, Feiran Huang, and Xiao Huang. 2025.
\newblock Next-generation database interfaces: A survey of llm-based text-to-sql.
\newblock \emph{IEEE Transactions on Knowledge and Data Engineering}.

\bibitem[{Huang et~al.(2024)Huang, Ma, Li, Huang, Zou, Zhang, and Zheng}]{DBLP:conf/coling/HuangMLHZ0Z24}
Shulin Huang, Shirong Ma, Yinghui Li, Mengzuo Huang, Wuhe Zou, Weidong Zhang, and Haitao Zheng. 2024.
\newblock \href {https://aclanthology.org/2024.lrec-main.889} {Lateval: An interactive llms evaluation benchmark with incomplete information from lateral thinking puzzles}.
\newblock In \emph{Proceedings of the 2024 Joint International Conference on Computational Linguistics, Language Resources and Evaluation, {LREC/COLING} 2024, 20-25 May, 2024, Torino, Italy}, pages 10186--10197. {ELRA} and {ICCL}.

\bibitem[{Hui et~al.(2024)Hui, Yang, Cui, Yang, Liu, Zhang, Liu, Zhang, Yu, Dang et~al.}]{qwen25coder}
Binyuan Hui, Jian Yang, Zeyu Cui, Jiaxi Yang, Dayiheng Liu, Lei Zhang, Tianyu Liu, Jiajun Zhang, Bowen Yu, Kai Dang, and 1 others. 2024.
\newblock Qwen2.5-coder technical report.
\newblock \emph{arXiv preprint arXiv:2409.12186}.

\bibitem[{Hurst et~al.(2024)Hurst, Lerer, Goucher, Perelman, Ramesh, Clark, Ostrow, Welihinda, Hayes, Radford et~al.}]{gpt4o}
Aaron Hurst, Adam Lerer, Adam~P Goucher, Adam Perelman, Aditya Ramesh, Aidan Clark, AJ~Ostrow, Akila Welihinda, Alan Hayes, Alec Radford, and 1 others. 2024.
\newblock Gpt-4o system card.
\newblock \emph{arXiv preprint arXiv:2410.21276}.

\bibitem[{Iyyer et~al.(2017)Iyyer, Yih, and Chang}]{iyyer2017search}
Mohit Iyyer, Wen-tau Yih, and Ming-Wei Chang. 2017.
\newblock Search-based neural structured learning for sequential question answering.
\newblock In \emph{ACL 2017}, pages 1821--1831.

\bibitem[{Jaech et~al.(2024)Jaech, Kalai, Lerer, Richardson, El-Kishky, Low, Helyar, Madry, Beutel, Carney et~al.}]{o1}
Aaron Jaech, Adam Kalai, Adam Lerer, Adam Richardson, Ahmed El-Kishky, Aiden Low, Alec Helyar, Aleksander Madry, Alex Beutel, Alex Carney, and 1 others. 2024.
\newblock Openai o1 system card.
\newblock \emph{arXiv preprint arXiv:2412.16720}.

\bibitem[{Jiang et~al.(2026)Jiang, Dong, Kang, Zhu, and Song}]{jiang2026foefores}
Kehan Jiang, Haonan Dong, Zhaolu Kang, Zhengzhou Zhu, and Guojie Song. 2026.
\newblock \href {https://arxiv.org/abs/2604.02967} {Foe: Forest of errors makes the first solution the best in large reasoning models}.
\newblock \emph{Preprint}, arXiv:2604.02967.

\bibitem[{Jin et~al.(2022)Jin, Siebert, Li, and Chen}]{jin2022survey}
Nengzheng Jin, Joanna Siebert, Dongfang Li, and Qingcai Chen. 2022.
\newblock A survey on table question answering: recent advances.
\newblock In \emph{China Conference on Knowledge Graph and Semantic Computing}, pages 174--186. Springer.

\bibitem[{Li et~al.(2023)Li, Xu, Chen, Huang, Li, Jiang, Li, Zhou, Zheng, and Shen}]{DBLP:journals/corr/abs-2311-11268}
Yinghui Li, Zishan Xu, Shaoshen Chen, Haojing Huang, Yangning Li, Yong Jiang, Zhongli Li, Qingyu Zhou, Hai{-}Tao Zheng, and Ying Shen. 2023.
\newblock \href {https://doi.org/10.48550/ARXIV.2311.11268} {Towards real-world writing assistance: {A} chinese character checking benchmark with faked and misspelled characters}.
\newblock \emph{CoRR}, abs/2311.11268.

\bibitem[{Li et~al.(2024{\natexlab{a}})Li, Zhou, Luo, Ma, Li, Zheng, Hu, and Yu}]{DBLP:journals/corr/abs-2402-11100}
Yinghui Li, Qingyu Zhou, Yuanzhen Luo, Shirong Ma, Yangning Li, Hai{-}Tao Zheng, Xuming Hu, and Philip~S. Yu. 2024{\natexlab{a}}.
\newblock \href {https://doi.org/10.48550/ARXIV.2402.11100} {When llms meet cunning questions: {A} fallacy understanding benchmark for large language models}.
\newblock \emph{CoRR}, abs/2402.11100.

\bibitem[{Li et~al.(2024{\natexlab{b}})Li, Wu, Li, He, Fang, Zhang, Zhao, Li, Li, and Song}]{DBLP:conf/semweb/LiWLHFZZLLS24}
Zhongqiu Li, Zhenhe Wu, Mengxiang Li, Zhongjiang He, Ruiyu Fang, Jie Zhang, Yu~Zhao, Yongxiang Li, Zhoujun Li, and Shuangyong Song. 2024{\natexlab{b}}.
\newblock Scalable database-driven kgs can help text-to-sql.
\newblock In \emph{Proceedings of the {ISWC} 2024 Posters, Demos and Industry Tracks: From Novel Ideas to Industrial Practice co-located with 23nd International Semantic Web Conference {(ISWC} 2024), Hanover, Maryland, USA, November 11-15, 2024}, {CEUR} Workshop Proceedings. CEUR-WS.org.

\bibitem[{Lin(2004)}]{rouge}
Chin-Yew Lin. 2004.
\newblock {ROUGE}: A package for automatic evaluation of summaries.
\newblock In \emph{Text Summarization Branches Out}, Barcelona, Spain. Association for Computational Linguistics.

\bibitem[{Lin et~al.(2024)Lin, Ruan, Liu, and Wang}]{DBLP:journals/tkde/LinRLW24}
Yupian Lin, Tong Ruan, Jingping Liu, and Haofen Wang. 2024.
\newblock A survey on neural data-to-text generation.
\newblock \emph{{IEEE} Trans. Knowl. Data Eng.}, 36(4):1431--1449.

\bibitem[{Liu et~al.(2025{\natexlab{a}})Liu, Wang, Yang, Jiang, Zhao, Wang, Li, He, Liu, Yuan, Gao, Wang, Yao, Xiong, Deng, He, Yu, Zhao, Fang, Jiang, Li, Hu, Yu, Li, Liu, Li, Shi, Niu, Huang, Xiao, Wang, Li, Pu, Jia, Yao, Huang, He, Jiang, Song, Xue, Xie, Zhang, Huang, Zhang, Lu, Zhang, Zhang, Xue, Yuan, Su, Jiang, Song, Li, and Li}]{telechat3}
Xinzhang Liu, Chao Wang, Zhihao Yang, Zhuo Jiang, Xudong Zhao, Haoran Wang, Lei Li, Dongdong He, Luobin Liu, Kaizhe Yuan, Han Gao, Zihan Wang, Yitong Yao, Sishi Xiong, Wenmin Deng, Haowei He, Kaidong Yu, Yu~Zhao, Ruiyu Fang, and 35 others. 2025{\natexlab{a}}.
\newblock Training report of telechat3-moe.
\newblock \emph{CoRR}, abs/2512.24157.

\bibitem[{Liu et~al.(2025{\natexlab{b}})Liu, Liu, Wen, Cai, Cui, He, and Zhang}]{liu2025uniform}
Zheng Liu, Mengjie Liu, Siwei Wen, Mengzhang Cai, Bin Cui, Conghui He, and Wentao Zhang. 2025{\natexlab{b}}.
\newblock From uniform to heterogeneous: Tailoring policy optimization to every token's nature.
\newblock \emph{arXiv preprint arXiv:2509.16591}.

\bibitem[{Mueller et~al.(2019)Mueller, Piccinno, Shaw, Nicosia, and Altun}]{mueller2019answering}
Thomas Mueller, Francesco Piccinno, Peter Shaw, Massimo Nicosia, and Yasemin Altun. 2019.
\newblock Answering conversational questions on structured data without logical forms.
\newblock In \emph{EMNLP-IJCNLP 2019}, pages 5902--5910.

\bibitem[{Nahid and Rafiei(2024)}]{DBLP:conf/naacl/NahidR24}
Md~Mahadi~Hasan Nahid and Davood Rafiei. 2024.
\newblock Tabsqlify: Enhancing reasoning capabilities of llms through table decomposition.
\newblock In \emph{Proceedings of the 2024 Conference of the North American Chapter of the Association for Computational Linguistics: Human Language Technologies (Volume 1: Long Papers), {NAACL} 2024, Mexico City, Mexico, June 16-21, 2024}, pages 5725--5737. Association for Computational Linguistics.

\bibitem[{Nakamura et~al.(2022)Nakamura, Levy, Tuan, Chen, and Wang}]{DBLP:conf/acl/NakamuraLTCW22}
Kai Nakamura, Sharon Levy, Yi{-}Lin Tuan, Wenhu Chen, and William~Yang Wang. 2022.
\newblock Hybridialogue: An information-seeking dialogue dataset grounded on tabular and textual data.
\newblock In \emph{Findings of the Association for Computational Linguistics: {ACL} 2022, Dublin, Ireland, May 22-27, 2022}, pages 481--492. Association for Computational Linguistics.

\bibitem[{Nan et~al.(2022)Nan, Hsieh, Mao, Lin, Verma, Zhang, Kry{\'s}ci{\'n}ski, Schoelkopf, Kong, Tang et~al.}]{nan2022fetaqa}
Linyong Nan, Chiachun Hsieh, Ziming Mao, Xi~Victoria Lin, Neha Verma, Rui Zhang, Wojciech Kry{\'s}ci{\'n}ski, Hailey Schoelkopf, Riley Kong, Xiangru Tang, and 1 others. 2022.
\newblock Fetaqa: Free-form table question answering.
\newblock \emph{TACL 2022}, 10:35--49.

\bibitem[{OpenAI(2023)}]{gpt4}
OpenAI. 2023.
\newblock \href {https://arxiv.org/abs/2303.08774} {Gpt-4 technical report}.
\newblock \emph{arXiv preprint arXiv:2303.08774}.

\bibitem[{OpenAI(2025)}]{gpt45}
OpenAI. 2025.
\newblock \href {https://openai.com/index/introducing-gpt-4-5/} {Introducing gpt-4.5}.

\bibitem[{Parikh et~al.(2020)Parikh, Wang, Gehrmann, Faruqui, Dhingra, Yang, and Das}]{parikh2020totto}
Ankur Parikh, Xuezhi Wang, Sebastian Gehrmann, Manaal Faruqui, Bhuwan Dhingra, Diyi Yang, and Dipanjan Das. 2020.
\newblock Totto: A controlled table-to-text generation dataset.
\newblock In \emph{EMNLP 2020}, pages 1173--1186.

\bibitem[{Pasupat and Liang(2015{\natexlab{a}})}]{wikitq}
Panupong Pasupat and Percy Liang. 2015{\natexlab{a}}.
\newblock Compositional semantic parsing on semi-structured tables.
\newblock In \emph{Proceedings of the 53rd Annual Meeting of the Association for Computational Linguistics and the 7th International Joint Conference on Natural Language Processing of the Asian Federation of Natural Language Processing, {ACL} 2015, July 26-31, 2015, Beijing, China, Volume 1: Long Papers}.

\bibitem[{Pasupat and Liang(2015{\natexlab{b}})}]{pasupat2015compositional}
Panupong Pasupat and Percy Liang. 2015{\natexlab{b}}.
\newblock Compositional semantic parsing on semi-structured tables.
\newblock In \emph{ACL 2015}, pages 1470--1480.

\bibitem[{Redmon and Farhadi(2017)}]{DBLP:conf/cvpr/RedmonF17}
Joseph Redmon and Ali Farhadi. 2017.
\newblock {YOLO9000:} better, faster, stronger.
\newblock In \emph{2017 {IEEE} Conference on Computer Vision and Pattern Recognition, {CVPR} 2017, Honolulu, HI, USA, July 21-26, 2017}, pages 6517--6525. {IEEE} Computer Society.

\bibitem[{Roziere et~al.(2023)Roziere, Gehring, Gloeckle, Sootla, Gat, Tan, Adi, Liu, Remez, Rapin et~al.}]{codellama}
Baptiste Roziere, Jonas Gehring, Fabian Gloeckle, Sten Sootla, Itai Gat, Xiaoqing~Ellen Tan, Yossi Adi, Jingyu Liu, Tal Remez, J{\'e}r{\'e}my Rapin, and 1 others. 2023.
\newblock Code llama: Open foundation models for code.
\newblock \emph{CoRR}.

\bibitem[{Schulman et~al.(2017)Schulman, Wolski, Dhariwal, Radford, and Klimov}]{ppo}
John Schulman, Filip Wolski, Prafulla Dhariwal, Alec Radford, and Oleg Klimov. 2017.
\newblock Proximal policy optimization algorithms.
\newblock \emph{arXiv preprint arXiv:1707.06347}.

\bibitem[{Shao et~al.(2024)Shao, Wang, Zhu, Xu, Song, Bi, Zhang, Zhang, Li, Wu et~al.}]{grpo}
Zhihong Shao, Peiyi Wang, Qihao Zhu, Runxin Xu, Junxiao Song, Xiao Bi, Haowei Zhang, Mingchuan Zhang, YK~Li, Y~Wu, and 1 others. 2024.
\newblock Deepseekmath: Pushing the limits of mathematical reasoning in open language models, 2024.
\newblock \emph{URL https://arxiv. org/abs/2402.03300}.

\bibitem[{Su et~al.(2024)Su, Wang, Ye, Zhou, Zhang, Zhu, Wang, Xu, Chen, Li et~al.}]{tablegpt2}
Aofeng Su, Aowen Wang, Chao Ye, Chen Zhou, Ga~Zhang, Guangcheng Zhu, Haobo Wang, Haokai Xu, Hao Chen, Haoze Li, and 1 others. 2024.
\newblock Tablegpt2: A large multimodal model with tabular data integration.
\newblock \emph{arXiv preprint arXiv:2411.02059}.

\bibitem[{Sui et~al.(2024)Sui, Zhou, Zhou, Han, and Zhang}]{table_meet_llm}
Yuan Sui, Mengyu Zhou, Mingjie Zhou, Shi Han, and Dongmei Zhang. 2024.
\newblock Table meets llm: Can large language models understand structured table data? a benchmark and empirical study.
\newblock In \emph{Proceedings of the 17th ACM International Conference on Web Search and Data Mining}, pages 645--654.

\bibitem[{Wang et~al.(2024{\natexlab{a}})Wang, Liu, Liu, Yao, Huang, Li, He, Li, Pu, Xu, Wang, and Song}]{wang-etal-2024-telechat}
Zihan Wang, XinZhang Liu, Shixuan Liu, Yitong Yao, Yunyao Huang, Mengxiang Li, Zhongjiang He, Yongxian Li, Luwen Pu, Huinan Xu, Chao Wang, and Shuangyong Song. 2024{\natexlab{a}}.
\newblock {T}ele{C}hat: An open-source billingual large language model.
\newblock In \emph{Proceedings of the 10th SIGHAN Workshop on Chinese Language Processing (SIGHAN-10)}.

\bibitem[{Wang et~al.(2024{\natexlab{b}})Wang, Liu, Liu, Yao, Huang, He, Li, Li, Che, Zhang, Wang, Wang, Pu, Xu, Fang, Zhao, Zhang, Huang, Lu, Peng, Zheng, Wang, Yang, He, Jiang, Xie, Zhang, Li, Shi, Fu, Zhang, Huang, Xiong, Zhang, Wang, and Song}]{telechat}
Zihan Wang, Xinzhang Liu, Shixuan Liu, Yitong Yao, Yuyao Huang, Zhongjiang He, Xuelong Li, Yongxiang Li, Zhonghao Che, Zhaoxi Zhang, Yan Wang, Xin Wang, Luwen Pu, Huihan Xu, Ruiyu Fang, Yu~Zhao, Jie Zhang, Xiaomeng Huang, Zhilong Lu, and 17 others. 2024{\natexlab{b}}.
\newblock Telechat technical report.
\newblock \emph{CoRR}, abs/2401.03804.

\bibitem[{Wang et~al.(2025)Wang, Liu, Yao, Wang, Zhao, Yang, Deng, Jia, Peng, Huang, Xiong, Jiang, Yu, Hu, Yao, Fang, Jiang, Song, Xie, Xue, He, Xue, Yuan, Zhang, Huang, Wang, Wang, Wu, Wang, Zhan, Sun, Xing, Jiang, Yang, Song, Li, He, and Li}]{telechat2}
Zihan Wang, Xinzhang Liu, Yitong Yao, Chao Wang, Yu~Zhao, Zhihao Yang, Wenmin Deng, Kaipeng Jia, Jiaxin Peng, Yuyao Huang, Sishi Xiong, Zhuo Jiang, Kaidong Yu, Xiaohui Hu, Fubei Yao, Ruiyu Fang, Zhuoru Jiang, Ruiting Song, Qiyi Xie, and 19 others. 2025.
\newblock Technical report of telechat2, telechat2.5 and {T1}.
\newblock \emph{CoRR}, abs/2507.18013.

\bibitem[{Wang et~al.(2024{\natexlab{c}})Wang, Zhang, Li, Eisenschlos, Perot, Wang, Miculicich, Fujii, Shang, Lee, and Pfister}]{DBLP:conf/iclr/0002ZLEP0MFSLP24}
Zilong Wang, Hao Zhang, Chun{-}Liang Li, Julian~Martin Eisenschlos, Vincent Perot, Zifeng Wang, Lesly Miculicich, Yasuhisa Fujii, Jingbo Shang, Chen{-}Yu Lee, and Tomas Pfister. 2024{\natexlab{c}}.
\newblock Chain-of-table: Evolving tables in the reasoning chain for table understanding.
\newblock In \emph{The Twelfth International Conference on Learning Representations, {ICLR} 2024, Vienna, Austria, May 7-11, 2024}. OpenReview.net.

\bibitem[{Wang et~al.(2024{\natexlab{d}})Wang, Zhang, Li, Eisenschlos, Perot, Wang, Miculicich, Fujii, Shang, Lee et~al.}]{chain_of_table}
Zilong Wang, Hao Zhang, Chun-Liang Li, Julian~Martin Eisenschlos, Vincent Perot, Zifeng Wang, Lesly Miculicich, Yasuhisa Fujii, Jingbo Shang, Chen-Yu Lee, and 1 others. 2024{\natexlab{d}}.
\newblock Chain-of-table: Evolving tables in the reasoning chain for table understanding.
\newblock \emph{arXiv preprint arXiv:2401.04398}.

\bibitem[{Wei et~al.(2022)Wei, Wang, Schuurmans, Bosma, Ichter, Xia, Chi, Le, and Zhou}]{cot}
Jason Wei, Xuezhi Wang, Dale Schuurmans, Maarten Bosma, Brian Ichter, Fei Xia, Ed~H. Chi, Quoc~V. Le, and Denny Zhou. 2022.
\newblock \href {http://papers.nips.cc/paper\_files/paper/2022/hash/9d5609613524ecf4f15af0f7b31abca4-Abstract-Conference.html} {Chain-of-thought prompting elicits reasoning in large language models}.
\newblock In \emph{Advances in Neural Information Processing Systems 35: Annual Conference on Neural Information Processing Systems 2022, NeurIPS 2022, New Orleans, LA, USA, November 28 - December 9, 2022}.

\bibitem[{Wu et~al.(2026)Wu, Zhang, Chang, Zhang, Liu, and Du}]{wu2026step}
Fei Wu, Zhenrong Zhang, Qikai Chang, Jianshu Zhang, Quan Liu, and Jun Du. 2026.
\newblock Step potential advantage estimation: Harnessing intermediate confidence and correctness for efficient mathematical reasoning.
\newblock \emph{arXiv preprint arXiv:2601.03823}.

\bibitem[{Wu et~al.(2024)Wu, Yang, Chai, Zhang, Liu, Du, Liang, Shu, Cheng, Sun et~al.}]{tablebench}
Xianjie Wu, Jian Yang, Linzheng Chai, Ge~Zhang, Jiaheng Liu, Xinrun Du, Di~Liang, Daixin Shu, Xianfu Cheng, Tianzhen Sun, and 1 others. 2024.
\newblock Tablebench: A comprehensive and complex benchmark for table question answering.
\newblock \emph{arXiv preprint arXiv:2408.09174}.

\bibitem[{Wu et~al.(2025{\natexlab{a}})Wu, Li, Li, Zhang, He, Yang, Zhao, Fang, Li, Li, and Song}]{MR-SQL}
Zhenhe Wu, Zhongqiu Li, Mengxiang Li, Jie Zhang, Zhongjiang He, Jian Yang, Yu~Zhao, Ruiyu Fang, Yongxiang Li, Zhoujun Li, and Shuangyong Song. 2025{\natexlab{a}}.
\newblock {MR-SQL:} multi-level retrieval enhances inference for {LLM} in text-to-sql.
\newblock In \emph{Database Systems for Advanced Applications - 30th International Conference, {DASFAA} 2025, Singapore, Singapore, May 26-29, 2025, Proceedings, Part {II}}, Lecture Notes in Computer Science, pages 403--413. Springer.

\bibitem[{Wu et~al.(2025{\natexlab{b}})Wu, Li, Zhang, He, Yang, Zhao, Fang, Wang, Xie, Song, and Li}]{UCS-SQL}
Zhenhe Wu, Zhongqiu Li, Jie Zhang, Zhongjiang He, Jian Yang, Yu~Zhao, Ruiyu Fang, Bing Wang, Hongyan Xie, Shuangyong Song, and Zhoujun Li. 2025{\natexlab{b}}.
\newblock {UCS-SQL:} uniting content and structure for enhanced semantic bridging in text-to-sql.
\newblock In \emph{Findings of the Association for Computational Linguistics, {ACL} 2025, Vienna, Austria, July 27 - August 1, 2025}, Findings of {ACL}, pages 8156--8168. Association for Computational Linguistics.

\bibitem[{Wu and Feng(2024)}]{wu-feng-2024-protrix}
Zirui Wu and Yansong Feng. 2024.
\newblock {P}ro{T}rix: Building models for planning and reasoning over tables with sentence context.
\newblock In \emph{Findings of the Association for Computational Linguistics: EMNLP 2024}.

\bibitem[{Yang et~al.(2025{\natexlab{a}})Yang, Li, Yang, Zhang, Hui, Zheng, Yu, Gao, Huang, Lv et~al.}]{qwen3}
An~Yang, Anfeng Li, Baosong Yang, Beichen Zhang, Binyuan Hui, Bo~Zheng, Bowen Yu, Chang Gao, Chengen Huang, Chenxu Lv, and 1 others. 2025{\natexlab{a}}.
\newblock \href {https://arxiv.org/abs/2505.09388} {Qwen3 technical report}.
\newblock \emph{Preprint}, arXiv:2505.09388.

\bibitem[{Yang et~al.(2024{\natexlab{a}})Yang, Yang, Hui, Zheng, Yu, Zhou, Li, Li, Liu, Huang et~al.}]{qwen2}
An~Yang, Baosong Yang, Binyuan Hui, Bo~Zheng, Bowen Yu, Chang Zhou, Chengpeng Li, Chengyuan Li, Dayiheng Liu, Fei Huang, and 1 others. 2024{\natexlab{a}}.
\newblock Qwen2 technical report.
\newblock \emph{arXiv preprint arXiv:2407.10671}.

\bibitem[{Yang et~al.(2025{\natexlab{b}})Yang, Yang, Zhang, Hui, Zheng, Yu, Li, Liu, Huang, Wei et~al.}]{qwen25}
An~Yang, Baosong Yang, Beichen Zhang, Binyuan Hui, Bo~Zheng, Bowen Yu, Chengyuan Li, Dayiheng Liu, Fei Huang, Haoran Wei, and 1 others. 2025{\natexlab{b}}.
\newblock Qwen2.5 technical report.
\newblock \emph{CoRR}, abs/2412.15115.

\bibitem[{Yang et~al.(2025{\natexlab{c}})Yang, Liu, Lv, Deng, Guo, Jing, Li, Liu, Luo, Luo et~al.}]{yang2025code}
Jian Yang, Xianglong Liu, Weifeng Lv, Ken Deng, Shawn Guo, Lin Jing, Yizhi Li, Shark Liu, Xianzhen Luo, Yuyu Luo, and 1 others. 2025{\natexlab{c}}.
\newblock From code foundation models to agents and applications: A comprehensive survey and practical guide to code intelligence.
\newblock \emph{arXiv preprint arXiv:2511.18538}.

\bibitem[{Yang et~al.(2020{\natexlab{a}})Yang, Ma, Zhang, Li, and Zhou}]{soft_template}
Jian Yang, Shuming Ma, Dongdong Zhang, Zhoujun Li, and Ming Zhou. 2020{\natexlab{a}}.
\newblock \href {https://doi.org/10.18653/V1/2020.ACL-MAIN.531} {Improving neural machine translation with soft template prediction}.
\newblock In \emph{Proceedings of the 58th Annual Meeting of the Association for Computational Linguistics, {ACL} 2020, Online, July 5-10, 2020}, pages 5979--5989. Association for Computational Linguistics.

\bibitem[{Yang et~al.(2020{\natexlab{b}})Yang, Ma, Zhang, Wu, Li, and Zhou}]{alm}
Jian Yang, Shuming Ma, Dongdong Zhang, Shuangzhi Wu, Zhoujun Li, and Ming Zhou. 2020{\natexlab{b}}.
\newblock \href {https://doi.org/10.1609/AAAI.V34I05.6480} {Alternating language modeling for cross-lingual pre-training}.
\newblock In \emph{The Thirty-Fourth {AAAI} Conference on Artificial Intelligence, {AAAI} 2020, The Thirty-Second Innovative Applications of Artificial Intelligence Conference, {IAAI} 2020, The Tenth {AAAI} Symposium on Educational Advances in Artificial Intelligence, {EAAI} 2020, New York, NY, USA, February 7-12, 2020}, pages 9386--9393. {AAAI} Press.

\bibitem[{Yang et~al.(2024{\natexlab{b}})Yang, Yang, Jin, Miao, Zhang, Yang, Cui, Zhang, Hui, and Lin}]{codearena}
Jian Yang, Jiaxi Yang, Ke~Jin, Yibo Miao, Lei Zhang, Liqun Yang, Zeyu Cui, Yichang Zhang, Binyuan Hui, and Junyang Lin. 2024{\natexlab{b}}.
\newblock Evaluating and aligning codellms on human preference.
\newblock \emph{arXiv preprint arXiv:2412.05210}.

\bibitem[{Yang et~al.(2025{\natexlab{d}})Yang, Yang, Zhang, Jin, Miao, Zhang, Yang, Cui, Zhang, Li, Hui, and Lin}]{DBLP:conf/emnlp/YangYZKMZYCZLHL25}
Jian Yang, Jiaxi Yang, Wei Zhang, Ke~Jin, Yibo Miao, Lei Zhang, Liqun Yang, Zeyu Cui, Yichang Zhang, Zhoujun Li, Binyuan Hui, and Junyang Lin. 2025{\natexlab{d}}.
\newblock \href {https://doi.org/10.18653/V1/2025.EMNLP-MAIN.489} {Codearena: Evaluating and aligning codellms on human preference}.
\newblock In \emph{Proceedings of the 2025 Conference on Empirical Methods in Natural Language Processing, {EMNLP} 2025, Suzhou, China, November 4-9, 2025}, pages 9672--9683. Association for Computational Linguistics.

\bibitem[{Yang et~al.(2024{\natexlab{c}})Yang, Zhang, Yang, Jin, Zhang, Peng, Deng, Miao, Liu, Cui et~al.}]{execrepobench}
Jian Yang, Jiajun Zhang, Jiaxi Yang, Ke~Jin, Lei Zhang, Qiyao Peng, Ken Deng, Yibo Miao, Tianyu Liu, Zeyu Cui, and 1 others. 2024{\natexlab{c}}.
\newblock Execrepobench: Multi-level executable code completion evaluation.
\newblock \emph{arXiv preprint arXiv:2412.11990}.

\bibitem[{Yang et~al.(2026{\natexlab{a}})Yang, Zhang, Guo, Ye, Jing, Liu, Li, Wu, Liu, Ma et~al.}]{yang2026iquest}
Jian Yang, Wei Zhang, Shawn Guo, Zhengmao Ye, Lin Jing, Shark Liu, Yizhi Li, Jiajun Wu, Cening Liu, X~Ma, and 1 others. 2026{\natexlab{a}}.
\newblock Iquest-coder-v1 technical report.
\newblock \emph{arXiv preprint arXiv:2603.16733}.

\bibitem[{Yang et~al.(2025{\natexlab{e}})Yang, Zhang, Li, Guo, Wang, Liu, Zhang, Wang, Li, Liu et~al.}]{yang2025codesimpleqa}
Jian Yang, Wei Zhang, Yizhi Li, Shawn Guo, Haowen Wang, Aishan Liu, Ge~Zhang, Zili Wang, Zhoujun Li, Xianglong Liu, and 1 others. 2025{\natexlab{e}}.
\newblock Codesimpleqa: Scaling factuality in code large language models.
\newblock \emph{arXiv preprint arXiv:2512.19424}.

\bibitem[{Yang et~al.(2025{\natexlab{f}})Yang, Zhang, Miao, Quan, Wu, Peng, Yang, Liu, Cui, Hui, and Lin}]{DBLP:conf/acl/YangZMQW0Y0CHL25}
Jian Yang, Wei Zhang, Yibo Miao, Shanghaoran Quan, Zhenhe Wu, Qiyao Peng, Liqun Yang, Tianyu Liu, Zeyu Cui, Binyuan Hui, and Junyang Lin. 2025{\natexlab{f}}.
\newblock \href {https://aclanthology.org/2025.acl-long.642/} {Qwen2.5-xcoder: Multi-agent collaboration for multilingual code instruction tuning}.
\newblock In \emph{Proceedings of the 63rd Annual Meeting of the Association for Computational Linguistics (Volume 1: Long Papers), {ACL} 2025, Vienna, Austria, July 27 - August 1, 2025}, pages 13121--13131. Association for Computational Linguistics.

\bibitem[{Yang et~al.(2026{\natexlab{b}})Yang, Zhang, Wu, Cheng, Guo, Wang, Gu, Du, Li, Xu et~al.}]{yang2026incoder}
Jian Yang, Wei Zhang, Jiajun Wu, Junhang Cheng, Shawn Guo, Haowen Wang, Weicheng Gu, Yaxin Du, Joseph Li, Fanglin Xu, and 1 others. 2026{\natexlab{b}}.
\newblock Incoder-32b: Code foundation model for industrial scenarios.
\newblock \emph{arXiv preprint arXiv:2603.16790}.

\bibitem[{Yang et~al.(2026{\natexlab{c}})Yang, Zhang, Wu, Cheng, Zheng, Xu, Gu, Jing, Du, Li et~al.}]{yang2026incoder_thinking}
Jian Yang, Wei Zhang, Jiajun Wu, Junhang Cheng, Tuney Zheng, Fanglin Xu, Weicheng Gu, Lin Jing, Yaxin Du, Joseph Li, and 1 others. 2026{\natexlab{c}}.
\newblock Incoder-32b-thinking: Industrial code world model for thinking.
\newblock \emph{arXiv preprint arXiv:2604.03144}.

\bibitem[{Yang et~al.(2025{\natexlab{g}})Yang, Zhang, Yang, Miao, Quan, Wu, Peng, Yang, Liu, Cui et~al.}]{xcoder}
Jian Yang, Wei Zhang, Jiaxi Yang, Yibo Miao, Shanghaoran Quan, Zhenhe Wu, Qiyao Peng, Liqun Yang, Tianyu Liu, Zeyu Cui, and 1 others. 2025{\natexlab{g}}.
\newblock Multi-agent collaboration for multilingual code instruction tuning.
\newblock \emph{arXiv preprint arXiv:2502.07487}.

\bibitem[{Ye et~al.(2025)Ye, Guo, Jin, Shen, Hou, Chen, Yang, and Jiang}]{ye2025tableqa}
Shenghao Ye, Yu~Guo, Dong Jin, Yikai Shen, Yunpeng Hou, Shuangwu Chen, Jian Yang, and Xiaofeng Jiang. 2025.
\newblock When tableqa meets noise: A dual denoising framework for complex questions and large-scale tables.
\newblock \emph{arXiv preprint arXiv:2509.17680}.

\bibitem[{Ye et~al.(2023)Ye, Hui, Yang, Li, Huang, and Li}]{DBLP:conf/sigir/YeHYLHL23}
Yunhu Ye, Binyuan Hui, Min Yang, Binhua Li, Fei Huang, and Yongbin Li. 2023.
\newblock Large language models are versatile decomposers: Decomposing evidence and questions for table-based reasoning.
\newblock In \emph{Proceedings of the 46th International {ACM} {SIGIR} Conference on Research and Development in Information Retrieval, {SIGIR} 2023, Taipei, Taiwan, July 23-27, 2023}, pages 174--184. {ACM}.

\bibitem[{Yu et~al.(2018)Yu, Zhang, Yang, Yasunaga, Wang, Li, Ma, Li, Yao, Roman et~al.}]{yu2018spider}
Tao Yu, Rui Zhang, Kai Yang, Michihiro Yasunaga, Dongxu Wang, Zifan Li, James Ma, Irene Li, Qingning Yao, Shanelle Roman, and 1 others. 2018.
\newblock Spider: A large-scale human-labeled dataset for complex and cross-domain semantic parsing and text-to-sql task.
\newblock In \emph{EMNLP 2018}, pages 3911--3921.

\bibitem[{Yuan et~al.(2025)Yuan, Chen, Hong, Zhang, Huang, Li, and Huang}]{yuan2025knapsack}
Zheng Yuan, Hao Chen, Zijin Hong, Qinggang Zhang, Feiran Huang, Qing Li, and Xiao Huang. 2025.
\newblock Knapsack optimization-based schema linking for llm-based text-to-sql generation.
\newblock \emph{arXiv preprint arXiv:2502.12911}.

\bibitem[{Zhang et~al.(2024{\natexlab{a}})Zhang, Chen, Dong, Chen, Huang, and Huang}]{zhang2024structure}
Qinggang Zhang, Hao Chen, Junnan Dong, Shengyuan Chen, Feiran Huang, and Xiao Huang. 2024{\natexlab{a}}.
\newblock Structure guided large language model for sql generation.
\newblock \emph{arXiv preprint arXiv:2402.13284}.

\bibitem[{Zhang et~al.(2024{\natexlab{b}})Zhang, Yue, Li, and Sun}]{DBLP:conf/naacl/ZhangYL024}
Tianshu Zhang, Xiang Yue, Yifei Li, and Huan Sun. 2024{\natexlab{b}}.
\newblock Tablellama: Towards open large generalist models for tables.
\newblock In \emph{Proceedings of the 2024 Conference of the North American Chapter of the Association for Computational Linguistics: Human Language Technologies (Volume 1: Long Papers), {NAACL} 2024, Mexico City, Mexico, June 16-21, 2024}, pages 6024--6044. Association for Computational Linguistics.

\bibitem[{Zhang et~al.(2024{\natexlab{c}})Zhang, Zhang, Ma, Li, Zhang, Li, Yao, Xu, Zhou, Zhang{-}Li, Yu, Zhao, Li, and Tang}]{tablellm}
Xiaokang Zhang, Jing Zhang, Zeyao Ma, Yang Li, Bohan Zhang, Guanlin Li, Zijun Yao, Kangli Xu, Jinchang Zhou, Daniel Zhang{-}Li, Jifan Yu, Shu Zhao, Juanzi Li, and Jie Tang. 2024{\natexlab{c}}.
\newblock Tablellm: Enabling tabular data manipulation by llms in real office usage scenarios.
\newblock \emph{CoRR}, abs/2403.19318.

\bibitem[{Zhao et~al.(2026)Zhao, Ren, Zou, Yang, Xu, Ge, Chen, Ma, Shi, Wang et~al.}]{zhao2026reinforced}
Ziqi Zhao, Zhaochun Ren, Jiahong Zou, Liu Yang, Zhiwei Xu, Xuri Ge, Zhumin Chen, Xinyu Ma, Daiting Shi, Shuaiqiang Wang, and 1 others. 2026.
\newblock Reinforced efficient reasoning via semantically diverse exploration.
\newblock \emph{arXiv preprint arXiv:2601.05053}.

\bibitem[{Zhong et~al.(2017)Zhong, Xiong, and Socher}]{zhong2017seq2sql}
Victor Zhong, Caiming Xiong, and Richard Socher. 2017.
\newblock Seq2sql: Generating structured queries from natural language using reinforcement learning.
\newblock \emph{CoRR}.

\bibitem[{Zhou et~al.(2026)Zhou, Li, Cheng, Fan, and Cheng}]{zhou2026look}
Yixiao Zhou, Yang Li, Dongzhou Cheng, Hehe Fan, and Yu~Cheng. 2026.
\newblock Look inward to explore outward: Learning temperature policy from llm internal states via hierarchical rl.
\newblock \emph{arXiv preprint arXiv:2602.13035}.

\end{thebibliography}

\appendix
\clearpage
\begin{figure*}[ht]
\centerline{\includegraphics[width=1.95\columnwidth]{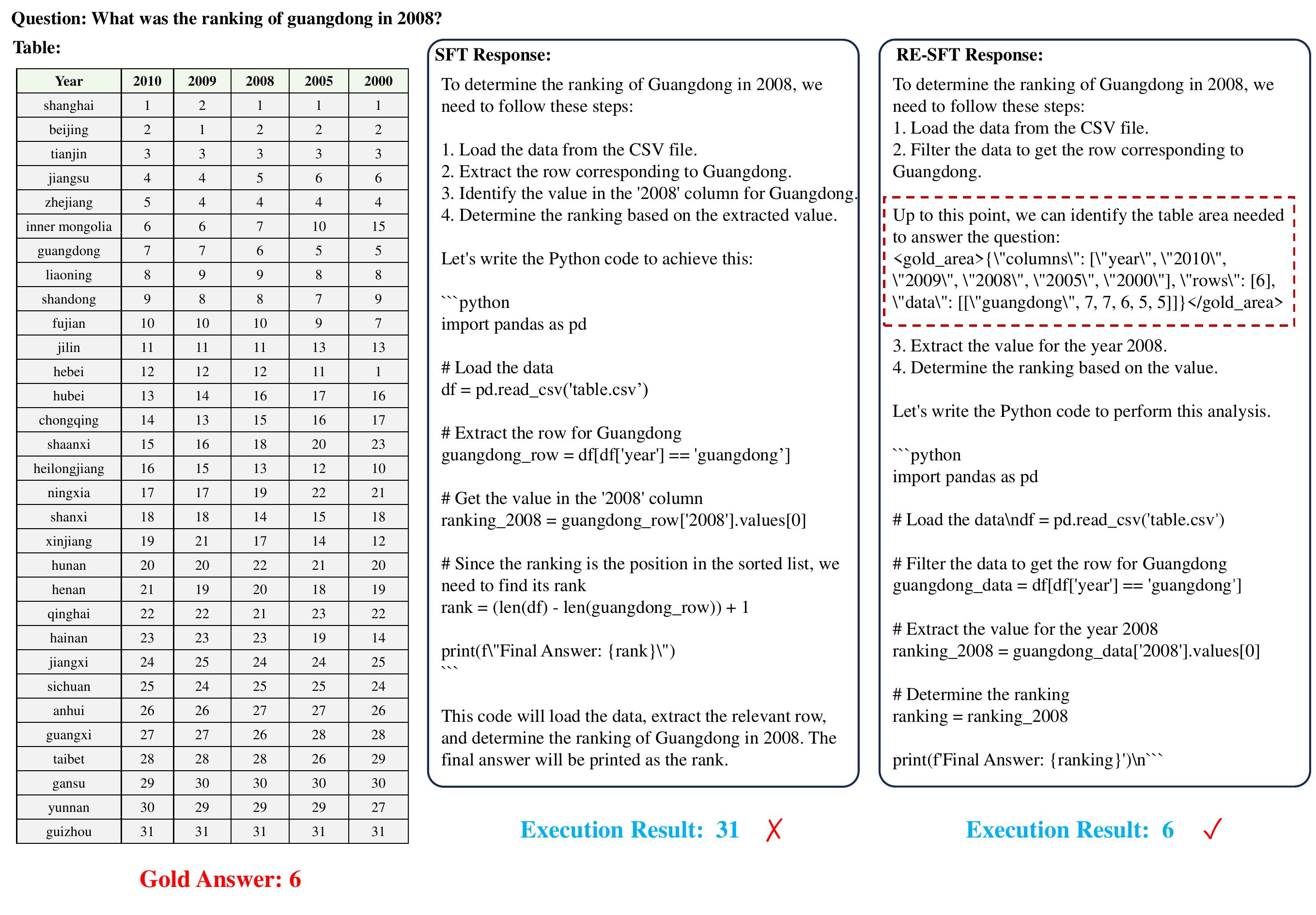}}
\caption{A case study for PoT, comparing SFT and RE-SFT.}
\label{POT_SFT}
\end{figure*}
\section{How RE-SFT enhances the performance of PoT}
\begin{table}[ht]
\begin{center}
\resizebox{0.5 \textwidth}{!}{
\begin{tabular}{cccccccc}
\bottomrule
\hline
\specialrule{0em}{1pt}{0pt}
\raisebox{-2pt}[0pt][0pt]{\multirow{2}{*} {Method}} & \raisebox{-2pt}[0pt][0pt]{\multirow{2}{*} {Base Model}}& \raisebox{-2pt}[0pt][0pt]{\multirow{2}{*} {Size}}& \raisebox{-2pt}[0pt][0pt]{\multirow{2}{*} {FC}}& \raisebox{-2pt}[0pt][0pt]{\multirow{2}{*} {NR}}& \raisebox{-2pt}[0pt][0pt]{\multirow{2}{*} {DA}}& \raisebox{-2pt}[0pt][0pt]{\multirow{2}{*} {VIZ}} & \raisebox{-2pt}[0pt][0pt]{\multirow{2}{*} {Overall}}\\\\
\hline
\specialrule{0em}{1pt}{0pt}
SFT &\multirow{2}{*} {CodeQwen}&\multirow{2}{*} {7B}& 11.46& 18.64& 14.28& 36.00& 15.14\\
RE-SFT &&&64.58&55.42&27.57&40.00&42.52\\
\hline
\specialrule{0em}{1pt}{0pt}
SFT &\multirow{2}{*} {DS-Coder}&\multirow{2}{*} {7B}& 10.42& 23.17& 18.58& 30.00& 18.72\\
RE-SFT &&&66.67& 55.16& 25.01& 36.00& 41.65\\
\hline
\specialrule{0em}{1pt}{0pt}
SFT &\multirow{2}{*} {Llama3}&\multirow{2}{*} {8B}& 12.50& 16.88& 15.02& 28.00& 14.75\\
RE-SFT &&&61.46& 53.40& 26.73& 28.00& 40.95\\
\hline
\specialrule{0em}{1pt}{0pt}
SFT &\multirow{2}{*} {Llama3.1}&\multirow{2}{*} {8B}& 27.08& 38.29& 18.16& 44.00& 27.14\\
RE-SFT &&&60.42& 55.92& 27.20& 32.00& 42.15\\
\hline
\specialrule{0em}{1pt}{0pt}
SFT &\multirow{2}{*} {Qwen2}&\multirow{2}{*} {7B}& 9.38& 13.35& 15.10& 26.00& 12.86\\
RE-SFT &&&61.46& 52.14& 18.08& 22.00& 41.07\\
\hline
\specialrule{0em}{1pt}{0pt}
SFT &\multirow{2}{*} {Qwen2.5}&\multirow{2}{*} {3B}& 38.54& 30.98& 24.48& 34.00& 27.55\\
RE-SFT &&&59.38& 52.90& 29.66& 32.00& 41.64\\
\hline
\specialrule{0em}{1pt}{0pt}
SFT &\multirow{2}{*} {Qwen3}&\multirow{2}{*} {8B}& 42.71& 33.50& 24.52& 42.00&  29.15\\
RE-SFT &&&67.71& 56.93& 29.70& 38.00& 44.36\\
\bottomrule
\hline
\end{tabular}}
\end{center}
\caption{Detailed results on PoT of TableBench. There are four types of questions: Fact Checking, Numerical Reasoning, Data Analysis, and Visualization.}
\label{pot}
\end{table}

In the main experiment results, RE-SFT significantly enhances the performance of PoT. Consequently, we conduct a detailed classification of questions to investigate the specific improvements in each category. As shown in Table~\ref{pot}, RE-SFT leads to substantial improvements in Fact Checking and Numerical Reasoning questions. It also results in partial enhancements in Data Analysis. However, it exerts no significant influence on Visualization.

We've conducted a case study on PoT data to investigate how model reasoning evolves after SFT and RE-SFT. Figure~\ref{POT_SFT} illustrates that RE-SFT model initially identifies a narrow table region, then generates code based on it, streamlining the code generation process. This approach reduces errors from ambiguous or incorrect reasoning compared to the SFT model.

\section{LLM Prompt Examples}
\begin{figure*}[ht]
\centerline{\includegraphics[width=1.9\columnwidth]{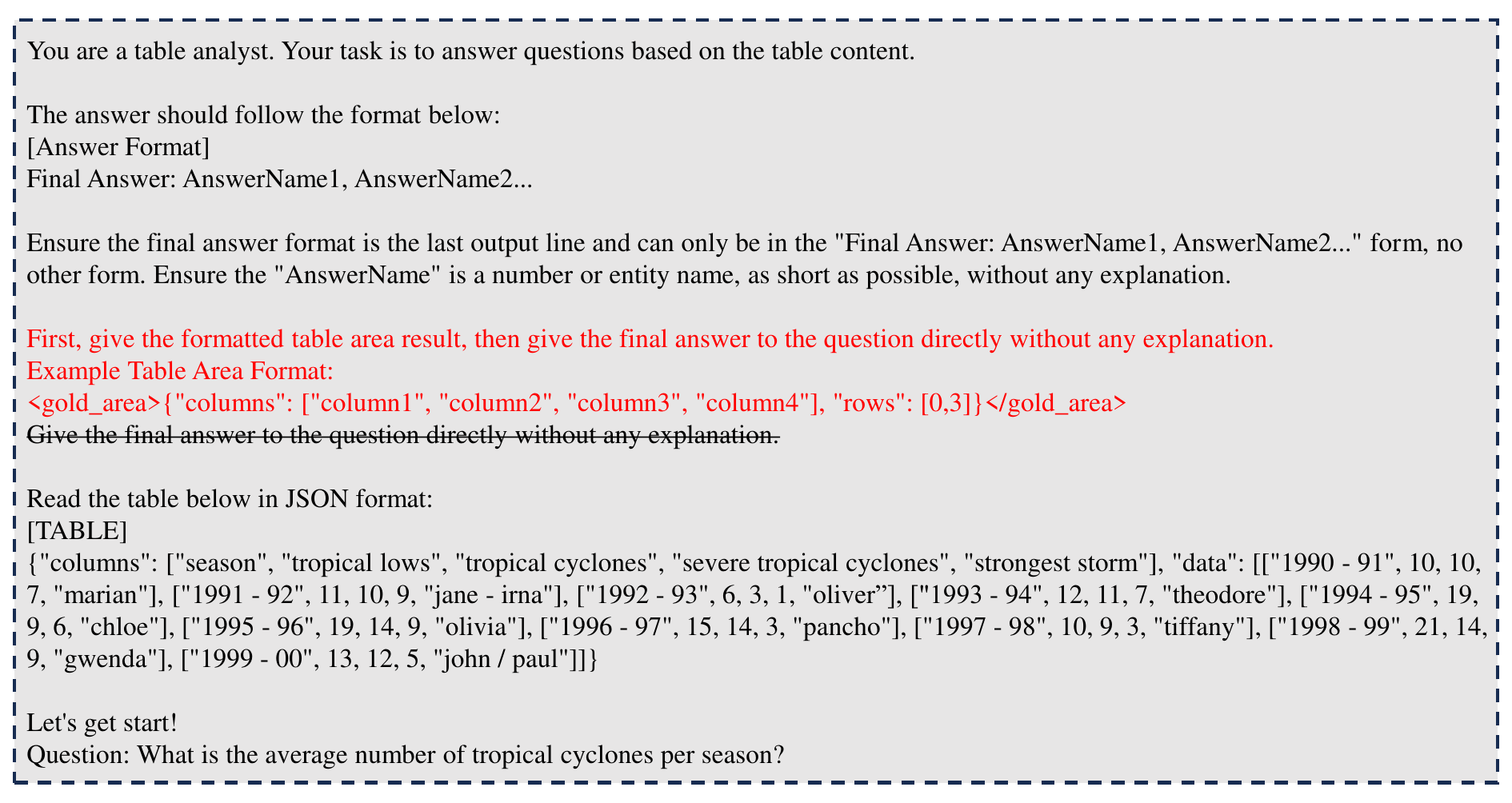}}
\caption{Instruction for DP data in TableBench.}
\label{DP}
\end{figure*}

\begin{figure*}[ht]
\centerline{\includegraphics[width=1.9\columnwidth]{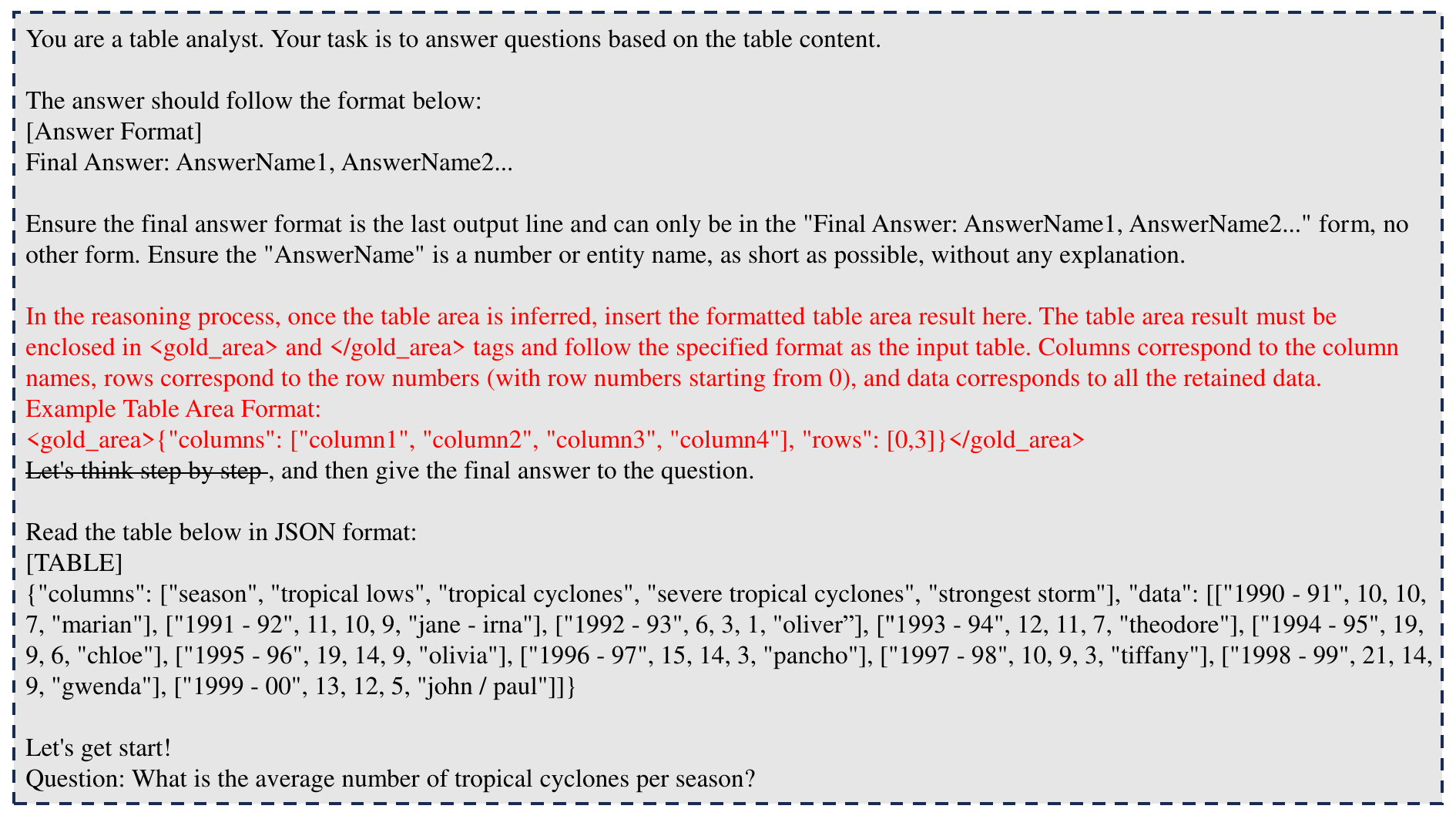}}
\caption{Instruction for TCoT data in TableBench (along with WikiTQ and WikiSQL in our experiments).}
\label{TCoT}
\end{figure*}

\begin{figure*}[ht]
\centerline{\includegraphics[width=1.9\columnwidth]{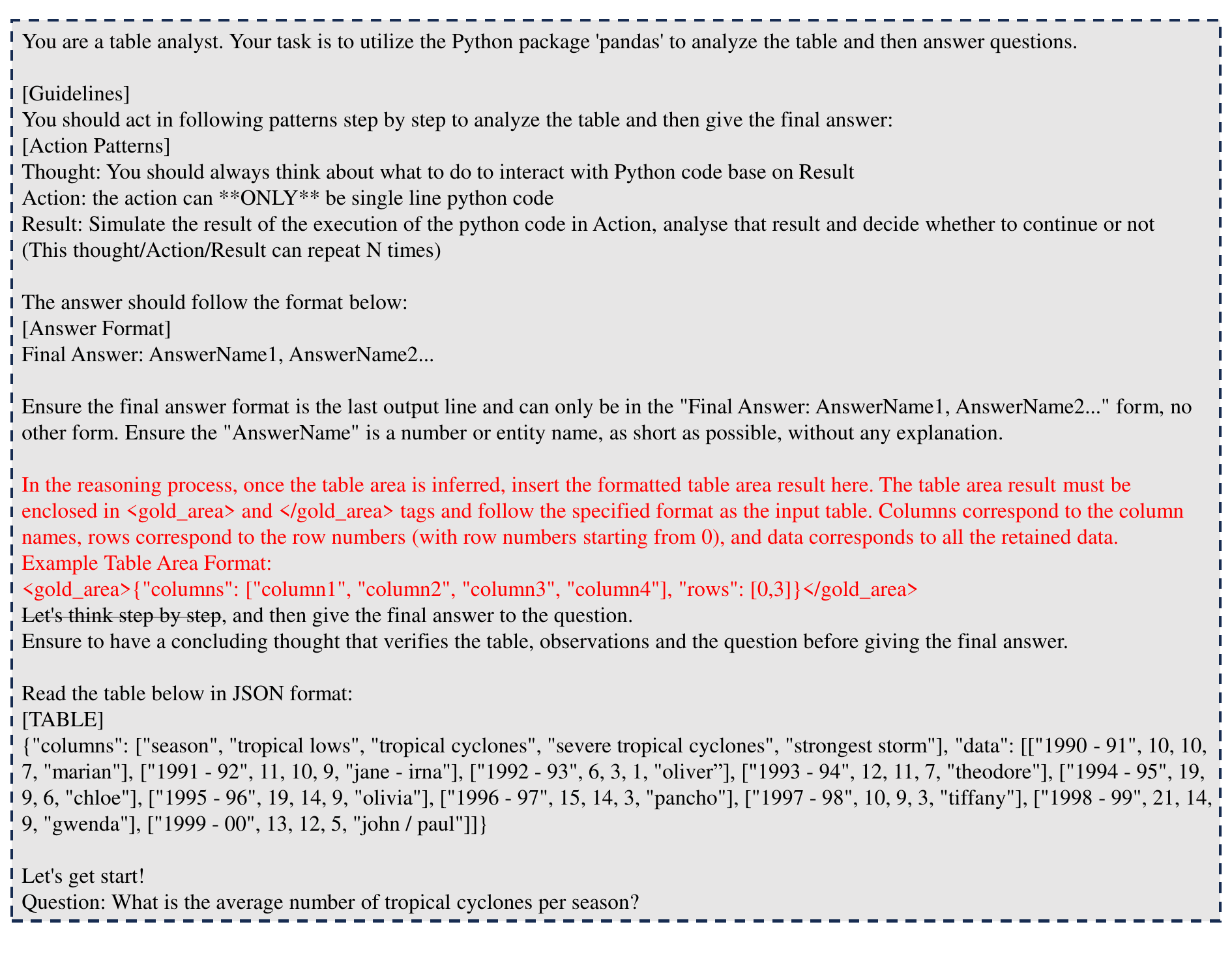}}
\caption{Instruction for SCoT data in TableBench.}
\label{SCoT}
\end{figure*}

\begin{figure*}[ht]
\centerline{\includegraphics[width=1.9\columnwidth]{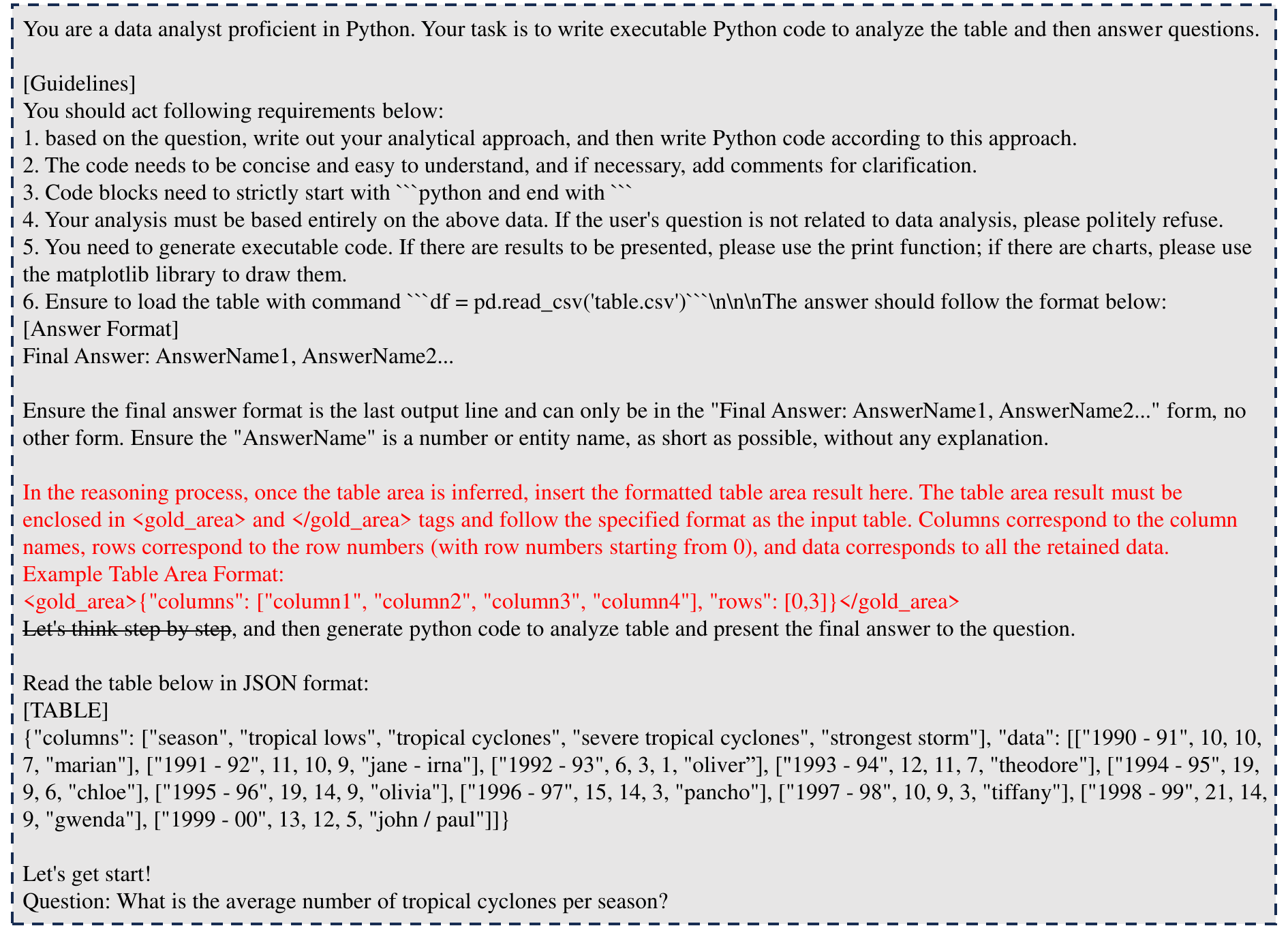}}
\caption{Instruction for PoT data in TableBench.}
\label{PoT}
\end{figure*}

\begin{figure*}[ht]
\centerline{\includegraphics[width=1.8\columnwidth]{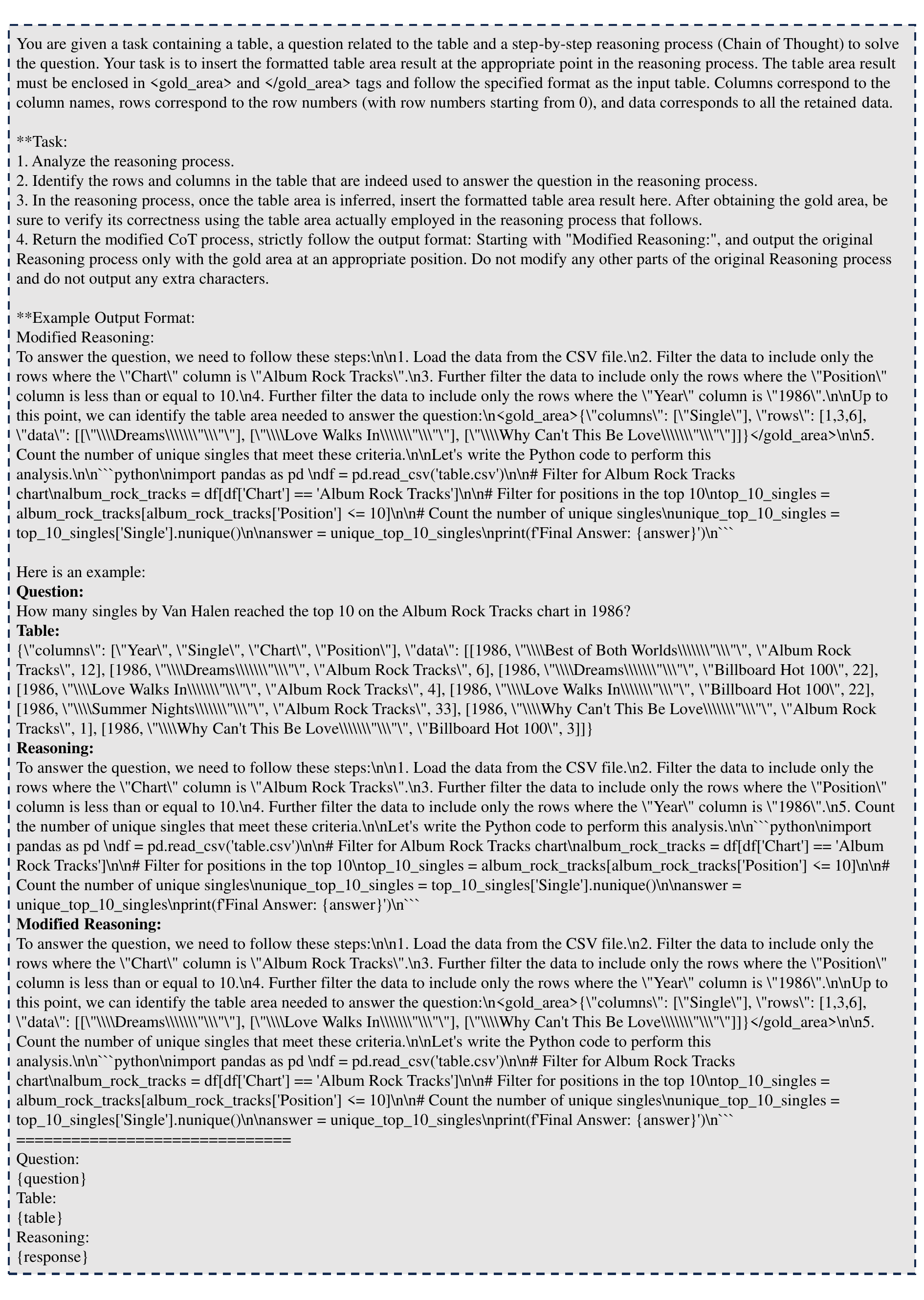}}
\caption{Instructions for DeepSeek-R1 inserting Table Regions in CoT.}
\label{DeepSeek-R1}
\end{figure*}

Figure~\ref{DP},~\ref{TCoT},~\ref{SCoT},~\ref{PoT} present the detailed prompts used to elicit the four distinct reasoning methods, while Figure~\ref{DeepSeek-R1} shows the instructions provided to DeepSeek-R1 for inserting table regions within the chain-of-thought.

\end{document}